%% file: neurips_2024.tex
\documentclass{article}

\usepackage[final]{neurips_2024}

\usepackage[colorlinks=true, linkcolor=red, urlcolor=black, citecolor=blue, anchorcolor=blue, backref=page]{hyperref}
\renewcommand*{\backref}[1]{}
\renewcommand*{\backrefalt}[4]{%
    \ifcase #1%
          \or [Cited on page~#2.]%
          \else [Cited on pages~#2.]%
    \fi%
    }
\usepackage{comment}
\usepackage{natbib}
\usepackage{times}
\usepackage{latexsym}
\usepackage{bm}

\usepackage[utf8]{inputenc} 
\usepackage[T1]{fontenc}    
\usepackage{hyperref}       
\usepackage{url}            
\usepackage{booktabs}       
\usepackage{amsfonts}       
\usepackage{nicefrac}       
\usepackage{microtype}      
\usepackage{xcolor}         
\usepackage{xspace}
\usepackage{pifont}
\usepackage{tikz}
\usetikzlibrary{bayesnet}
\usepackage{wrapfig}

\hypersetup{
    colorlinks=true,
    linkcolor=blue,
    }

\usepackage{inconsolata}
\usepackage{amsmath}
\usepackage{mathtools}
\usepackage{graphicx}
\usepackage{centernot}
\usepackage{amsfonts}
\usepackage{booktabs}
\newcommand{\indep}{\rotatebox[origin=c]{90}{$\models$}}
\usepackage{enumitem}
\usepackage{soul}
\usepackage{lineno}
\usepackage{algorithm}
\usepackage{algorithmic}
\usepackage{multirow}
\usepackage{wrapfig}

\usepackage{titletoc}

\newcommand\titleShort{\textbf{COLD}}
\newcommand\titleFull{\textbf{Causal reasOning in cLosed Daily activities}}

\usepackage{cleveref}
\usepackage{bbm}
\crefname{figure}{Figure}{Figures}
\crefname{table}{Table}{Tables}
\crefname{appendix}{Appendix}{Appendices}
\crefname{section}{Section}{Sections}
\crefname{equation}{Eq.}{Eqs.}
\crefname{enumi}{}{} %

\newcommand{\cmark}{\ding{51}}%
\newcommand{\xmark}{\ding{55}}%

\title{COLD: Causal reasOning in cLosed Daily activities}

\author{%
  Abhinav Joshi\thanks{Equal Contribution} \qquad Areeb Ahmad\footnotemark[1]\qquad Ashutosh Modi \\
  Department of Computer Science and Engineering\\
  Indian Institute of Technology Kanpur (IIT Kanpur)\\
  Kanpur, India \\
  \texttt{\{ajoshi,areeb,ashutoshm\}@cse.iitk.ac.in} \\
}

\begin{document}

\maketitle


\input{./sections/abstract}

\input{./sections/introduction}
\input{./sections/background}

\input{./sections/methodology}
\input{./sections/experiments}

\input{./sections/relatedwork}
\input{./sections/limitations}
\input{./sections/conclusion}

\section*{Acknowledgments}

We would like to thank the anonymous reviewers and the meta-reviewer for their insightful comments and suggestions. 
We would like to thank Google Deepmind India for helping us with the conference travel support. 


\bibliography{references}

\bibliographystyle{plainnat}

\newpage
\appendix
\input{./sections/appendix}



\clearpage
\newpage

\end{document}

%% file: sections/abstract.tex
\begin{abstract}
Large Language Models (LLMs) have shown state-of-the-art performance in a variety of tasks, including arithmetic and reasoning; however, to gauge the intellectual capabilities of LLMs, causal reasoning has become a reliable proxy for validating a general understanding of the mechanics and intricacies of the world similar to humans. Previous works in natural language processing (NLP) have either focused on open-ended causal reasoning via causal commonsense reasoning (CCR) or framed a symbolic representation-based question answering for theoretically backed-up analysis via a causal inference engine. The former adds an advantage of real-world grounding but lacks theoretically backed-up analysis/validation, whereas the latter is far from real-world grounding. In this work, we bridge this gap by proposing the \titleShort\ (\titleFull) framework, which is built upon human understanding of daily real-world activities to reason about the causal nature of events. We show that the proposed framework facilitates the creation of enormous causal queries ($\sim 9$ million) and comes close to the mini-turing test, simulating causal reasoning to evaluate the understanding of a daily real-world task. We evaluate multiple LLMs on the created causal queries and find that causal reasoning is challenging even for activities trivial to humans. We further explore (the causal reasoning abilities of LLMs) using the backdoor criterion to determine the causal strength between events. 
\end{abstract}

%% file: sections/introduction.tex
\section{Introduction} \label{sec:intro}

In recent times, Large Language Models (LLMs) have shown remarkable generalization capabilities \citep{devlin-etal-2019-bert, Radford2019LanguageMA, incontextfewshotlearners}. Consequently, the ability to perform causal reasoning (often considered a core feature of intelligence \citep{causal_cognition,bookOfWhy}) has sparked research interest in the context of LLMs, aiming to answer if causal reasoning is possible with LLMs \citep{weber2020causal,jin2023largecorr2cause,jin2024cladder,cohrs2023large,romanou-etal-2023-crab,yang2023a,mitchell2023comparing,vashishtha2023causal,stolfo-etal-2023-causal}. 
On a broader level, there are two lines of work;  
first, that treats the causal reasoning via learning relationships between the events that are grounded in the real world \citep{gordon-etal-2012-semeval_copa_dataset,ho2022wikiwhy,2023causalparrots,zhang2023understanding, wang-etal-2023-cola}. Second line of work relies on a causal inference engine and establishes relationships between variables via symbolic representation \citep{jin2023largecorr2cause, jin2024cladder}. 
The former relies on understanding real-world events but lacks formal definitions that adhere to the causal inference theory. 
The latter solves the issue using a causal inference engine but uses symbolic representations not grounded in the world, 
making the causal queries more like a test for the understanding of causal theory. 
Though the first line of work includes real-world events, the causal queries are often limited and could be answered by memorizing the causal relationships between the events. Recent findings that include rigorous analysis using a causal inference engine claim LLMs to be \textit{``Causal Parrots''} \citep{2023causalparrots}, i.e., the LLMs tend to pick up (memorize) patterns in the training data to perform well on the causal reasoning benchmarks. 
Moreover, some initial findings by  \cite{tang2023large} suggest that LLMs perform significantly better when semantics are consistent with commonsense but struggle to solve symbolic tasks, pointing towards semantic representation to be better for proper validation of LLMs, leading to a conclusion that an in-depth analysis using real-world events is necessary.
%

In this work, we bridge the gap between the two approaches by proposing the \titleShort\ (\titleFull) framework, based on the human understanding of real-world daily activities capturing commonsense (for example, ``making coffee,'' ``boarding an airplane,'' etc.), that adheres to causal theory literature. It is more natural to frame real-life reasoning-based queries via language; consequently, we follow the literature on Causal Commonsense Reasoning (CCR), which studies the relationships between real-world events (described via natural language). 
\begin{wrapfigure}{r}{0.5\textwidth}
\centering
 \includegraphics[width=0.48\textwidth]{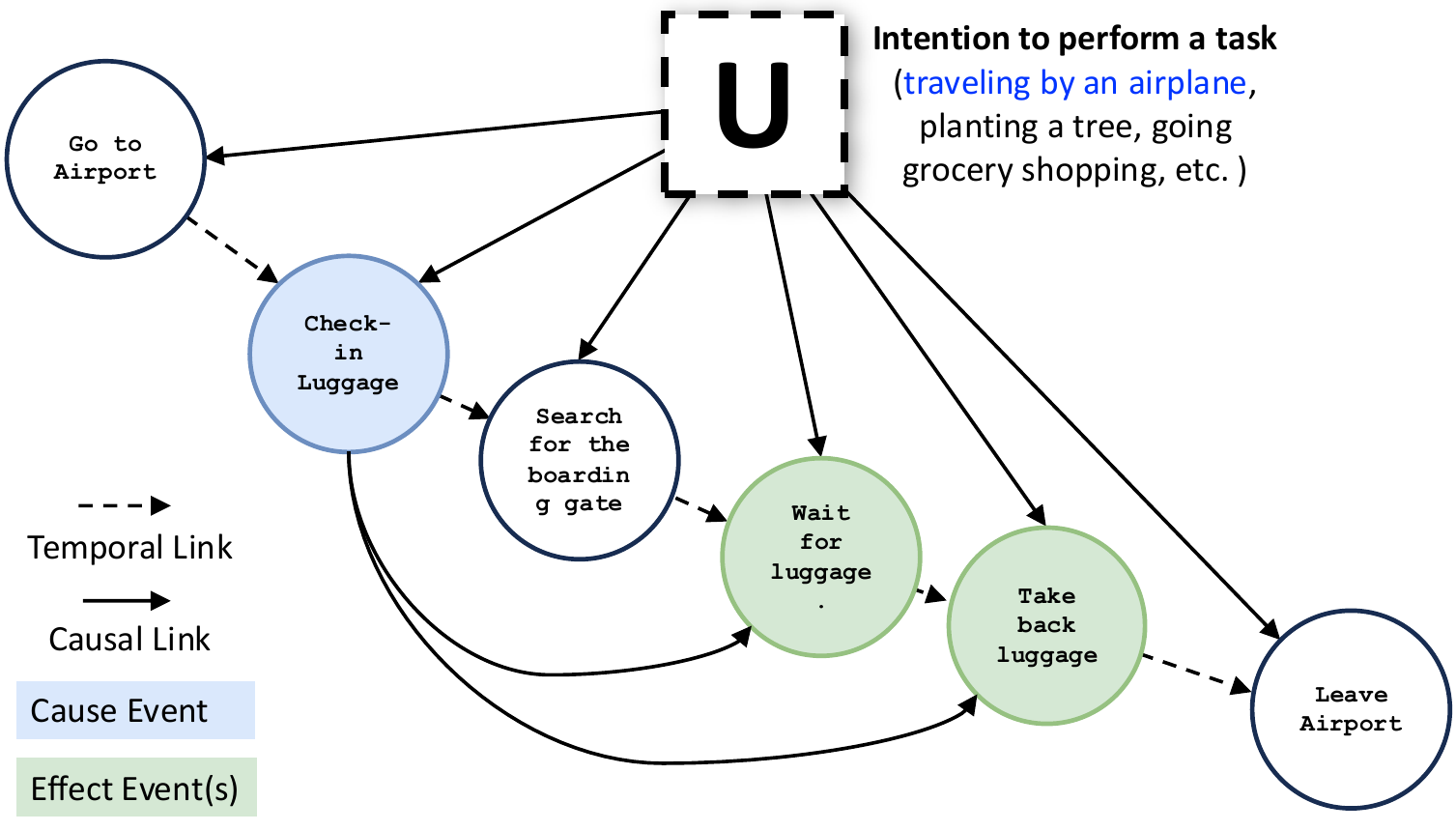}
  \caption{
  \textbf{U} denotes the unobserved variables, confounding all events present in a real-world activity. In an activity, some events cause other events to happen. For example, in \textit{``traveling by an airplane''}, the event of \textit{``check-in luggage''} causes events like \texttt{``taking back luggage.''}}
  \label{fig:causal_thumbnail_intro}
\end{wrapfigure}
CCR is a non-trivial task of estimating the cause-and-effect relationship between events that are studied under the umbrella of commonsense reasoning \citep{KUIPERS1984169,gordon-etal-2012-semeval_copa_dataset,pmlr-v162-zhang22am-rock,wang-etal-2023-cola, chun-etal-2023-cretihc, du-etal-2022-e}.  
The events in CCR generally refer to actions taking place in an activity in the real world. For example, consider the activity of \texttt{``traveling by an airplane''} given in Fig.  \ref{fig:causal_thumbnail_intro}, where the occurrence of all the events is confounded by a universal variable $\mathbf{U}$ (``intention to perform a task''). Moreover, there are a few events that cause one another. For example, the event \texttt{``checking in luggage''} ($E_1$) caused the occurrence of events like \texttt{``waiting at the luggage belt''} ($E_2$) after the flight, i.e., in an alternate universe where one does not checks in luggage and goes with the cabin bags, will never wait for their luggage after the flight has landed. Moreover, some of the events have no causal impact, like \texttt{``find the boarding gate''} ($E_3$) has no causal relationship with \texttt{``checking in luggage''} ($E_1$). More formally, 
\begin{align}
\begin{split}
    \Delta_{(E_1 \rightarrow E_2)} = \mathbb{P}(E_2|do(E_1)) - \mathbb{P}(E_2|do(\neg E_1)) \\
    \Delta_{(E_1 \rightarrow E_3)} = \mathbb{P}(E_3|do(E_1)) - \mathbb{P}(E_3|do(\neg E_1))
\end{split}
\label{eq:average_treatment_effect}
\end{align}
where $do(.)$ denotes the \textbf{do} operator \citep{do_calculus} showing the intervention on $E_1$, and $\Delta$ is the \textit{causal estimand} capturing the causal strength between two events, i.e., $\Delta (E_1 \rightarrow E_2)$ is expected to be higher when compared to $\Delta (E_1 \rightarrow E_3)$. Note, CCR excludes the causal events that are beyond the reach of commonsense knowledge, for example, does \texttt{``planting trees''} have a direct impact on the \texttt{``rainy season''}? or does \texttt{``providing free education''} improve the \texttt{``economic condition of the country/state''}; does \texttt{``carpooling''} directly impact \texttt{``air pollution''}, etc. A noteworthy point concerning causality is that though the logical temporal (or prototypical) order of these events provides a weak signal about causal relationships, temporal precedence does not always imply causation (\S\ref{sec:preliminaries}). For example, one could erroneously argue that \texttt{``boarding a plane''} is also the cause of \texttt{``waiting at the luggage belt''} since without \texttt{``boarding a plane,''} one cannot wait for the luggage belt. 

\begin{figure*}[t]
\centering
\includegraphics[width=\linewidth]{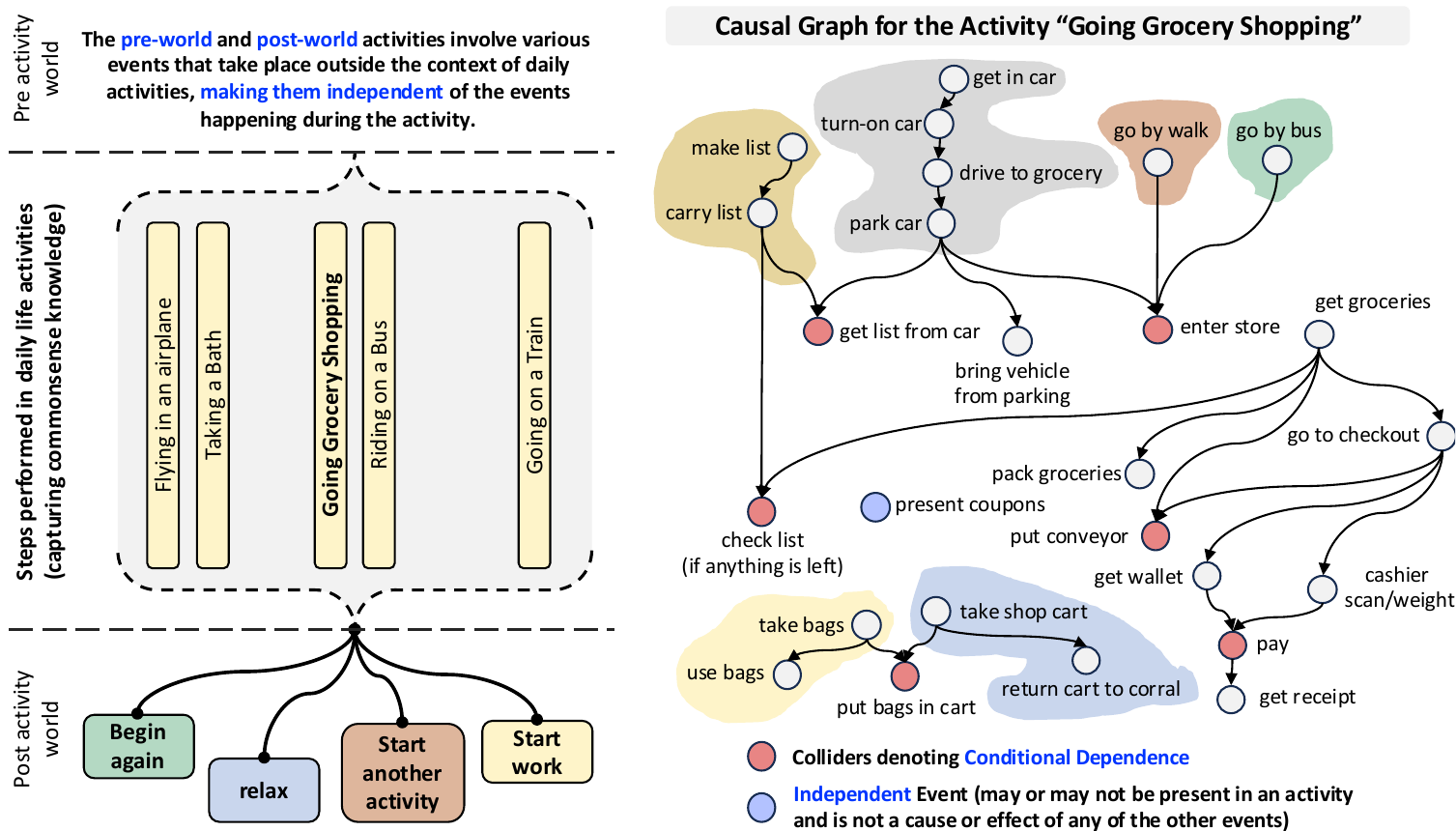} 
\caption{
\textbf{Left:} the figure represents the closed nature of daily real-world activities (capturing commonsense, commonly understood by humans), start and end given the context of the task, i.e., the pre-activity world and post-activity world activities marginalize out the dependence of event occurring during the activity with the rest of the world. \textbf{Right:} Causal Graph for \texttt{``going grocery shopping.''} Notice the collider (red nodes) makes the independent set of nodes (highlighted in different colors) unconditionally independent in the causal graph. In contrast, when given a condition on a collider (\texttt{``put bags in cart''}, the two clusters (yellow and blue) become dependent (if collider is observed, both yellow and blue clusters may have been observed as well).}
\label{fig:causal_thumbnail_intro_graph_left_right} %
\vspace{-5mm} 
\end{figure*}

For building a causal reasoning framework (based on CCR) around real-life daily activities, one would require a few primary features readily available: 1) Clear distinction between the events, i.e., the events should encapsulate describing a particular step in an activity, 2) Causal Dependency between the variables/events, i.e., there should be some events causing other events to occur. 3) Causal independence of events with the rest of the world, i.e., the occurrence of events should be independent of events that are not part of the activity (i.e., the covariates are balanced out). We found that \textit{``Scripts''} \citep{SCHANK1975237, schank1975scripts} provide a concrete medium that satisfies all these requirements.  Scripts are defined as a sequence of events describing a prototypical activity, such as going to a restaurant, and hence capture commonsense knowledge about the world. \citep{schank1975scripts, modi-etal-2016-inscript, wanzare-etal-2016-crowdsourced, ostermann-etal-2018-mcscript, modi-2016-event,Modi17-Thesis, modi-etal-2017-modeling, modi2014inducing}. Moreover, different people have similar understandings of the activity in the form of scripts that inherently balance out the covariates present in the real world, i.e., all the activities have the same starting and ending point and account for common exogenous and endogenous variables, providing a suitable platform to establish a cause-and-effect relationship between the events. 
In other words, for an activity like \texttt{``flying in an airplane,''} or \texttt{``going grocery shopping''} (also see Fig.  \ref{fig:causal_thumbnail_intro_graph_left_right}, left) the events that happened before starting the activity and after completing the activity are marginalized out using a common understanding of these activities by different humans and hence will have no causal relations with any of the exogenous events during the activity. Creating a causal graph for script knowledge, i.e., establishing relationships between events taking place during the activity, provides a perfect platform for creating causal queries, thus providing a medium to establish CCR between events. In a nutshell, we make the following contributions:
\begin{itemize}[noitemsep,nosep,leftmargin=*]
\item We propose \titleShort\ (\titleFull), a CCR
framework based on Script knowledge (daily activities involving commonsense) that provides a closed system to test the understanding of causal inference grounded in the real-world. The proposed framework adheres to SUTVA (Stable Unit Treatment Value Assumption) \citep{Cox1958-COXPOE, Rubi:80} by design (\S\ref{sec:methods}). \titleShort\ consists of activity-specific observational graphs (created via crowd-sourcing) and causal graphs. Further, \titleShort\ facilitates creating an enormous number of causal queries (e.g., $2,887,950$ per activity) via causal query triplets from the causal graph. This comes close to the mini-Turing test \citep{bookOfWhy}, where the story becomes the understanding of the daily activity, and the sampled enormous causal queries help in the exhaustive and rigorous evaluation of LMs.
\item We devise various design mechanisms for estimating causal strength analytically and show how the representations learned by language models can be validated.
\item Via detailed experimentation on the widely used open-weight language models, including encoder-only models (RoBERTa-MNLI) and autoregressive models (gpt-neo-125M,
gpt-neo-1.3B, 
gemma-2b, 
gpt-neo-2.7B, 
phi-2, 
gpt-j-6B,
Llama-2-7b-chat-hf, 
Mistral-7B-v0.1,
gemma-7b, and
Meta-Llama-3-8B) 
we estimate the causal reasoning capability of the learned representations. 
We release the framework, model code, and results via 
\href{https://github.com/Exploration-Lab/COLD}{\color{blue}{https://github.com/Exploration-Lab/COLD}}.
\end{itemize}

%% file: sections/background.tex
\section{Background} \label{sec:preliminaries}

\textbf{The Mini Turing Test} proposed by \citet{bookOfWhy} is designed in a question-answering format to validate the understanding of causal knowledge about a simple story. 
The primary feature of a mini-Turing test is the enormous number of causal queries that can be framed using the underlying causal graph, which governs the occurrence of events in the story.
Due to the enormous number of causal queries, passing the mini-Turing via memorization becomes combinatorially heavy, and hence, the authors argue that it can only be beaten if one has access to the underlying causal graph governing the occurrence of events (i.e., one has the ability to reason causally about the events). 
In this work, though, we only consider a more straightforward case of choice-based causal triplets; we realize the number of causal queries that can be created is enormous and helps validate the causal reasoning abilities coming close to the mini-Turing test. 

\textbf{$\textit{d}$-separation:}
Establishing the independence of variables becomes non-trivial when dealing with complex interactions among multiple variables. 
{$\textit{d}$-separation} \citep{pearl1988probabilistic} facilitates the determination of conditional independence between two sets of nodes $\mathbf{X}$ and $\mathbf{Y}$ in a graphical model $\mathcal{{G}}$ given another set of nodes $\mathbf{Z}$.
$\textit{d}$-separation asserts that $\mathbf{X}$ and $\mathbf{Y}$, given the set $\textbf{Z}$, are $d$-separated if all paths for every 
node in $\mathbf{X}$ and every node in $\mathbf{Y}$ are blocked by conditioning on $\textbf{Z}$, denoted as $\textbf{X} \indep_{\mathcal{G}} \textbf{Y} \mid \textbf{Z}$. A path $p$ is blocked by a set of nodes $\mathbf{Z}$ \citep{pearl2016causal}, if and only if: 1) \textit{p} contains a chain of nodes $\mathbf{A}\rightarrow \mathbf{B}\rightarrow \mathbf{C}$ or a fork $\mathbf{A}\leftarrow \mathbf{B} \rightarrow \mathbf{C}$ such that the middle node $\mathbf{B}$ is in $\mathbf{Z}$  
OR 2) \textit{p} contains a collider $\mathbf{A} \rightarrow \mathbf{B}\leftarrow  \mathbf{C}$ such that the collision node $\mathbf{B}$ or its descendant is not in $\mathbf{Z}$.

\noindent\textbf{Backdoor Criterion:} 
A set of variables $W$ satisfies the backdoor criterion relative to $T$ and $Y$ if the following are true: 
\begin{itemize}[nosep,noitemsep]
    \item[(A)] $W$ blocks all backdoor paths from $T$ to $Y$ i.e. blocking confounding or non-causation association paths
    \item[(B)] $W$ doesn’t contain any descendants of $T$
\end{itemize}

Then, $W$ satisfies the backdoor criterion  \citep{pearl2016causal,neal2020introduction}. We make use of the backdoor criterion to estimate the causal estimand, capturing the relationship between the causal events. (Refer App. \ref{app-sec:backdoor} for more detail)

%% file: sections/methodology.tex
\begin{figure*}[t]
\centering
\includegraphics[width=\linewidth]{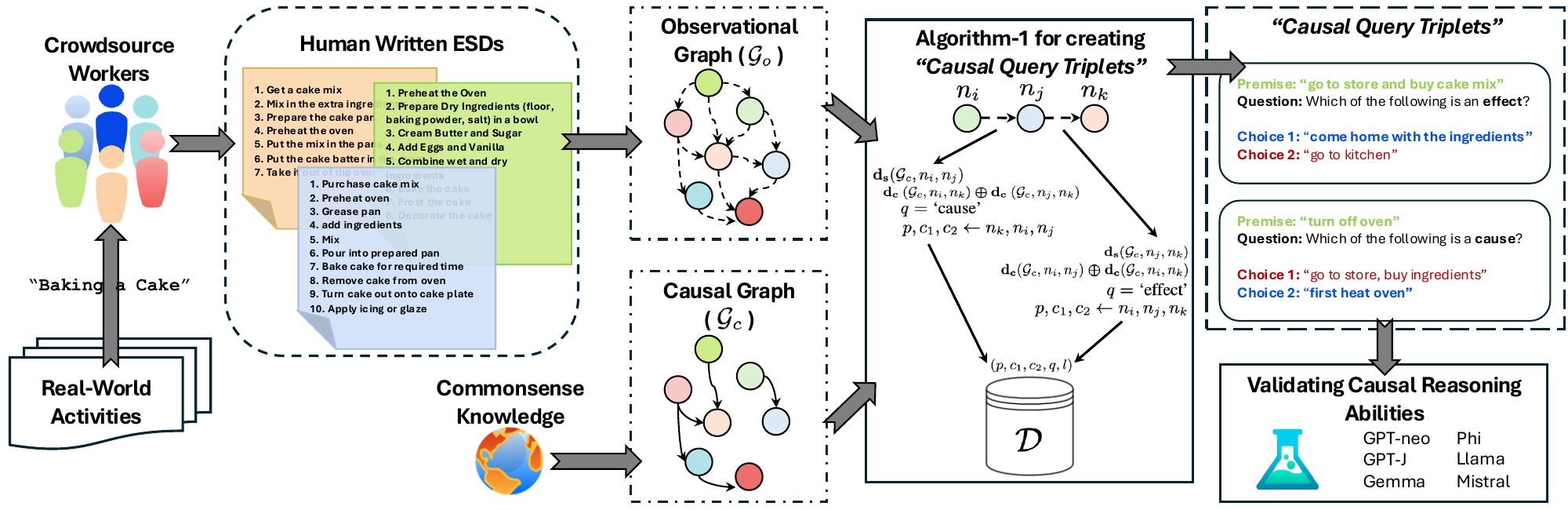} 
\caption{The proposed \titleShort\ framework for evaluating LLMs for causal reasoning.
The human-written Event Sequence Descriptions (ESDs) are obtained from crowdsource workers and include a telegrammic-style sequence of events when performing an activity. The Observational Graph and the Causal Graph for an activity are used to create causal query triplets (details in Algorithm \ref{alg:dataset_creation}), shown towards the right. 
Using counterfactual reasoning, “going to the kitchen” is possible without going to the market (if the ingredients are already available), making “come home with the ingredients.” a more plausible effect among the given choices.
Similarly, in the second example, the event “going to market” has no direct relation with the event “heating the oven”.
}
\label{fig:framework} %
\vspace{-5mm} 
\end{figure*}


\section{\titleShort\ (\titleFull)} \label{sec:methods}

We propose \titleShort\ (\titleFull) framework for testing causal reasoning abilities of natural language understanding systems such as LLMs. Fig. \ref{fig:framework} gives an overview of the creation process. We use crowd-sourced data of script knowledge to create observational graphs which is further used along with manual intervention to create causal graphs. Subsequently, an algorithm is used to create an enormous number of causal queries (causal triplets), which are further used to test LLMs for causal reasoning. Next, we explain each of the steps in more detail.

\noindent \textbf{Task Formulation:} \titleShort\ is motivated by Causal Commonsense Reasoning (CCR), which we define as the task of finding the strength of the cause-and-effect relationship between two events ($E_1$ and $E_2$) given in an activity $ a \in \mathcal{A}$, where $\mathcal{A}$ is the set of all activities. For example, for an activity like \texttt{``going in an airplane''}, the central question is to determine the causal relationship between two events that occur during the activity (events like \texttt{``checking in luggage''} and \texttt{``waiting for luggage''}). Since reasoning about a sequence of events is tedious (and sometimes confusing \citep{do-etal-2011-minimally}), researchers often rely on a more plausible cause rather than defining a definite causal event. For instance, COPA dataset \citep{gordon-etal-2012-semeval_copa_dataset} provides a premise event and a corresponding causal query question along with two choices (see Table \ref{tab:dataset:eg} for an example); a system is required to predict which of the two choices is most plausible cause/effect as required by the question.

\noindent\textbf{Creating a Closed Causal System}

Given the nature of Script knowledge (satisfying the criterion of balanced covariates, \S\ref{sec:intro}), we use a script corpus called DeScript \citep{wanzare-etal-2016-crowdsourced} for creating the observational graphs.  
DeScript is a corpus with a telegram-style sequential description of an activity in English (e.g., baking a cake, taking a bath, etc.). DeScript is created via crowd-sourcing. For a given activity, crowd-workers write a point-wise and sequential short description of various events involved in executing the activity (this one complete description is called an ESD (Event Sequence Description)).  DeScript collects data for a set of 40 daily activities (100 ESDs each) varying in complexity and background knowledge. %
Additionally, for a given activity, semantically similar events from different ESDs are manually aligned by human annotators (for more details, refer to \cite{wanzare-etal-2016-crowdsourced}). 
These alignments were later used by \citet{joshi2023scriptworld,joshi2023scripts-aamas} to create a DAG representing the overall activity.  
In our work, we use these DAGs as the observational distribution of an activity ($\mathcal{G}_o^{(a)}$, where $ a\in \mathcal{A}$, where $\mathcal{A}$ is the set of all activities). These DAGs provide a medium for generating enormous trajectories (scales from $1.6e+{16}$ to $1.3e+{27}$, also see Table \ref{tab:graph_details_and_results}), that are coming directly from human annotations (alignment as well as the ESDs), providing us a proxy to represent the understanding of daily activities.

\noindent \textbf{Observational Distribution ($\mathcal{G}_o$)}: Note that the graphs $\mathcal{G}_o$, approximately represent (almost) all possible ways in which an ESD can be written for an activity, providing the true observational distribution, i.e., how the combinations of events will look like while performing the activity in the real world (see App. \ref{app-sec:observation-graph} for examples).

\noindent \textbf{Causal Graphs ($\mathcal{G}_c$):} To reason about the causal relationships between the events (nodes of $\mathcal{G}_o$), we would need the underlying causal graph that shows the cause of occurrence of various activities (directly or indirectly). We construct the causal graphs manually by reasoning about the independence of various events in the activity. Fig. \ref{fig:causal_thumbnail_intro_graph_left_right} shows the pictorial representation of one of the created causal graphs for the activity \texttt{``going grocery shopping.''} Notice that various sets of events in the graph create clusters, denoting independence between various events. For example, nodes related to \texttt{make list} cause the events that involve the presence of a list and do not cause events like \texttt{going via car} (as some of the population will not create a list for shopping). Similarly, the mode of transportation (car/bus/walk) is independent of the events performed inside the store.  

\noindent\textbf{Causal Query Triplets:} 
The obtained Causal Graph ($\mathcal{G}_c$), for an activity provides a medium to reason about causal links between the events. 
Notice in Fig.  \ref{fig:causal_thumbnail_intro_graph_left_right}, how the red nodes (colliders) help separate out the independent event clusters. For example, the nodes `\texttt{get list from car}' and `\texttt{check list (if anything is left)}' being colliders, separate out the making list-related events with the rest of the graph. Similarly, the node `\texttt{put bags in cart}' separates out the `\texttt{take shop cart}' and `\texttt{take bags}'. Another interesting property represented in the obtained causal graph is the conditional dependence between various clusters. For example, the cluster related to `\texttt{get in car}' is unconditionally independent of `\texttt{make list}'. However, if we condition on the collider (`\texttt{get list from car}'), they become dependent, i.e., if `\texttt{get list from car}' is observed, it means that the person will have created the list as well as went by car for the grocery shopping  (similarly for node `\texttt{put bags in cart}'). $\textit{d}$-separation (\S\ref{sec:preliminaries})  provides an easier way to establish conditional/unconditional independence between the set of nodes. For creating the dataset of causal queries (similar to other datasets like COPA \citep{gordon-etal-2012-semeval_copa_dataset}), we need a triplet of three events (premise $p$, Choice-1 $c_1$, Choice-2 $c_2$) associated with a question about `cause' or `effect' relationship, i.e., given the premise which of the two choices is the cause/effect? (Table \ref{tab:dataset:eg}). We call these triplets \textit{Causal Query Triplets} that are used to frame a causal query between the events. App. \ref{app-sec:algorithms}, Algorithm \ref{alg:dataset_creation} presents the mechanism to create a dataset of causal query triplets. We start by constructing the set of possible triplets, and sort the nodes in every triplet using the topological order preset in the observational graph (a DAG). Further, using $\textit{d}$-separation, we figure out the triplets that have one node $\textit{d}$-separated from the other nodes. The $\textit{d}$-separated node becomes the wrong choice. The premise and correct choice are determined from the remaining choices, leading to a `cause' and `effect' query based on the topological order, i.e., the event occurring before (temporally) in an activity becomes `cause.' The other event becomes the plausible `effect.' Note that the temporal precedence is generally assumed essential for defining causation, and it is one of the most important clues that is used to distinguish causal from other types of associations \citep{mill1898system, hill1965environment, pearl1995theory}. For our framework, we also consider the topologically sorted order obtained from the observational graphs and use the temporal order to define the causal query triplets, i.e., the cause events will always precede the effect events (see App. \ref{app-sec:algorithms}, Algorithm \ref{alg:dataset_creation}, where $\mathcal{G}_o$ helps determine the temporal order of events). Table \ref{tab:dataset:eg} shows a sample of the created dataset in comparison to the COPA dataset. Note that each node (premise, Choice-1, Choice-2) in $\mathcal{G}_c$ also has multiple texts (instances of text describing the same event) written by different crowd-sourced workers. We can further use these text instances to enrich the created dataset by considering all the available instances to create all possible combinations. This strategy increases the scale of the created dataset by a huge margin. Overall, we found a dataset created from triplets using Algorithm \ref{alg:dataset_creation} results in \textbf{4,392} tuples, which, after using the text instances, increases to \textbf{9,520,340}, bringing it close to `mini turning test'' \citep{bookOfWhy}.

\begin{table*}
    \caption{The table shows examples of causal query triplets created using the causal graphs ($\mathcal{G}_c$) and observational graphs ($\mathcal{G}_o$) in Algorithm \ref{alg:dataset_creation}. The top row is taken from the 
        COPA dataset \citep{gordon-etal-2012-semeval_copa_dataset} for the purpose of comparison. Note the examples in the table show the samples taken from the instance version.}
    \label{tab:dataset:eg}
    \centering
    \resizebox{\linewidth}{!}{%
        \begin{tabular}{llllcc}
        
        \toprule
        \toprule
        \textbf{Dataset} & \textbf{Premise} & \textbf{Choice-1 (1)} & \textbf{Choice-2 (2)} & \textbf{question} & \textbf{answer} \\
        \midrule
    
    \multirow{3}{*}{\rotatebox[origin=c]{90}{\textsc{\textbf{Copa}}}} &
    \texttt{The man turned on the faucet.} & \texttt{The toilet filled with water.} & \texttt{Water flowed from the spout.} & effect & 2 \\
     & \texttt{The girl found a bug in her cereal.} & \texttt{She poured milk in the bowl.} & \texttt{She lost her appetite.} & effect & 2 \\
     & \texttt{The hamburger meat browned.} & \texttt{The cook froze it.} & \texttt{The cook grilled it.} & cause & 2 \\ \midrule
    
\multirow{5}{*}{\rotatebox[origin=c]{90}
     {\textsc{\begin{tabular}{@{}c@{}}\titleShort \\ \texttt{(Cake)}  \end{tabular} }}} &

    \texttt{buy proper ingredients.} & \texttt{go home with ingredients.} & \texttt{wait for the timer to go off.} & effect & 1 \\
&  \texttt{measure ingredients in designated measuring cups.} & \texttt{whisk after each addition.stir to combine.} & \texttt{clean up the mess.} & effect & 1 \\

 & \texttt{bake until cake is ready.} & \texttt{set timer.} & \texttt{carefully remove cake from pan.} & cause & 1 \\
 & \texttt{turn off oven.} & \texttt{go to store, buy ingredients.} & \texttt{first heat oven.} & cause & 2 \\

 & \texttt{preheat oven to 350 degrees.} & \texttt{turn off oven.} & \texttt{prepare the microwave oven and utensils} & effect & 1 \\

\midrule

\multirow{5}{*}{\rotatebox[origin=c]{90}
     {\textsc{\begin{tabular}{@{}c@{}}\titleShort \\ \texttt{(Shopping)}  \end{tabular} }}} &

        \texttt{pay total.} & \texttt{get receipt.} & \texttt{place cart into cart corral.} & effect & 1 \\

&  \texttt{get the bill for groceries.} & \texttt{pay for the grocery.} & \texttt{return cart to store.} & cause & 1 \\

    & \texttt{pay for it .} & \texttt{start at the non-cold side of the store.} & \texttt{go to shelf and get the food items.} & cause & 2 \\

&  \texttt{go back to the car.put the bags in the car.} & \texttt{take your car and drive to grocery shop.} & \texttt{bring items to checkout.} & cause & 1 \\

  &   \texttt{take the full cart to the checkout lane.} & \texttt{watch prices as the checker scans.} & \texttt{go down aisles.} & effect & 1 \\

\midrule

\multirow{5}{*}{\rotatebox[origin=c]{90}
     {\textsc{\begin{tabular}{@{}c@{}}\titleShort \\ \texttt{(Train)}  \end{tabular} }}} &

        \texttt{check train schedules.} & \texttt{choose a destination.} & \texttt{go to the car.} & cause & 1 \\

& \texttt{you board your train and find your seat.} & \texttt{find your seat or compartment.} & \texttt{wait for train.} & effect & 1 \\

  &   \texttt{get off train.} & \texttt{go out of the station.} & \texttt{take all the luggage out of train.} & effect & 1 \\

& \texttt{go out of the station.} & \texttt{put carry on luggage in overhead bin.} & \texttt{get off at your correct stop.} & cause & 2 \\

  &   \texttt{arrive at destination.} & \texttt{when train reaches destination, exit train.} & \texttt{walk to the train platform.} & effect & 1 \\

\midrule

\multirow{5}{*}{\rotatebox[origin=c]{90}
     {\textsc{\begin{tabular}{@{}c@{}}\titleShort \\ \texttt{(Tree)}  \end{tabular} }}} &

        \texttt{go to garden center.} & \texttt{transport it home.} & \texttt{choose type of tree.} & effect & 1 \\

& \texttt{fill hole with dirt and fertilizer.} & \texttt{get tree.} & \texttt{dig hole big enough for tree to grow.} & cause & 2 \\

  &   \texttt{place the tree at the top of the hole.} & \texttt{cover the roots with dirt.} & \texttt{get a tree.} & effect & 1 \\

& \texttt{fill in dirt around the tree gently.} & \texttt{take it home.} & \texttt{place the tree in the hole.} & cause & 2 \\

  &   \texttt{place tree sapling into hole.} & \texttt{pack dirt back in.} & \texttt{find place for tree.} & effect & 1 \\


    \midrule 

    \multirow{5}{*}{\rotatebox[origin=c]{90}
     {\textsc{\begin{tabular}{@{}c@{}}\titleShort \\ \texttt{(Bus)}  \end{tabular} }}} &

         \texttt{pull signal for stop.} & \texttt{bus stops at destination.} & \texttt{stand up and go to door.} & effect & 1 \\

& \texttt{board the bus when it arrives.} & \texttt{while boarding, pay the driver the required fee} & \texttt{pull signal for stop.} & effect & 1 \\
   & \texttt{step on bus.} & \texttt{take available seat.} & \texttt{wait for the bus to arrive.} & effect & 1 \\

& \texttt{buy bus ticket.} & \texttt{when it arrives get on.} & \texttt{when your stop approaches, pull cord.} & cause & 1 \\

&  \texttt{find seat on bus and sit.} & \texttt{find out what bus to take.} & \texttt{the bus arrives at the departure station.} & cause & 2 \\


\bottomrule
\bottomrule 
        \end{tabular}
    }
\end{table*}

\begin{table*}[t]
\caption{The table provides details of the observational graph ($\mathcal{G}_o$) for 5 activities. The Causal Query Triplets represent the total number of triplets generated via Algorithm \ref{alg:dataset_creation}. The instance version shows the number of samples present in the instance version (including different text instances describing the same event) of the created dataset. Table \ref{tab:dataset:eg} shows a small sample taken for the 5 activities. Overall, the huge number of samples highlights the exhaustive nature of evaluation that can be done for LLMs.}
\label{tab:graph_details_and_results}
\centering
\small
\renewcommand{\arraystretch}{1.2}
\setlength\tabcolsep{3pt}
\resizebox{\linewidth}{!}{%
\begin{tabular}{lccccccc}
\toprule
\textbf{Activity}                           & \begin{tabular}[c]{@{}c@{}} \textbf{Nodes} \end{tabular} & \textbf{Compact Trajectories} & \textbf{Total Trajectories}  &  \textbf{Causal Query Triplets}   &   \textbf{Instance version (Num. samples)} \\
\midrule
\texttt{Baking a \textbf{Cake}}               & 28                                                        & 177030 & $1.3e+27$ & 864 & 
2887950
\\
\texttt{Riding on a \textbf{Bus}}                     & 20                                                        & 13945 & $1.3e+17$ & 334 & 
834046
\\

\texttt{Going Grocery \textbf{Shopping}}      & 33                                                         & 626096 & $3.1e+26$ 
& 1984 & 
3739184 \\
\texttt{Going on a \textbf{Train}}                      & 26                                                 &     133799   & $4.9e+22$ &950 & 
1213114
\\
\texttt{Planting a \textbf{Tree}}                      & 23                                                    &  4466    &  $1.6e+16$ & 260 & 
846046 \\

\midrule
Total Dataset Samples                      & -                                                         & - & - & \textbf{4392} 
& 
\textbf{9,520,340}
\\
\bottomrule
\end{tabular}}
\end{table*}

\noindent \textbf{Adherence to SUTVA:} In causal literature, a fundamentally acknowledged \emph{Stable Unit Treatment Value Assumption} (SUTVA) \citep{Cox1958-COXPOE, Rubi:80}) requires that for each unit (e.g., sequence of events), there is only one version of the non-treatment, i.e., for an event in the sequence, there lie only two versions occurring and not occurring. SUTVA plays a vital role in causal inference by ensuring that each unit's treatment assignment has a consistent impact, facilitating the accurate estimation of treatment effects. Our framework closely adheres to SUTVA assumptions (details in \ref{app-sec:sutva}).

\noindent\textbf{Comparison with Other Causal Datasets:} 
We briefly compare the created dataset with the existing set of causal reasoning datasets in App. Table \ref{tab:causalbenchmarks}. The created dataset serves as a middle ground, having both real-world groundings as well as an underlying causal graph to create an exhaustive set of causal queries.

%% file: sections/experiments.tex
\section{Experiments and Results} \label{sec:exp-results}

\titleShort\ provides a causal query dataset for evaluating LMs for causal understanding. In particular, we consider the ``Causal Query Triplets" (Table \ref{tab:graph_details_and_results}) coming from compact trajectories as a base and sample the instance version coming from the same skeleton. Since it is not possible to evaluate all the possible causal queries that could be created using our framework, we use $10K$ samples for each activity to report our findings. For a fair comparison between various models and better reproducibility, we freeze the sampled causal query triplets and compare the success rate over the frozen samples. We evaluate via two methods. First, as done in previous work \cite{jin2024cladder, jin2023largecorr2cause, chen2024causal}, we first experiment with various LLMs using a prompt-based evaluation scheme; second, we propose other mechanisms (based on causal theory, e.g., Average Treatment Effect) that could be used to perform an in-depth analysis of evaluating causal relationships between events. 

\noindent\textbf{Causal Reasoning Evaluation of LLMs via Prompts:} 
We start with the prompt-based evaluation of recent open-weight LLMs (
    gpt-neo-125M,
    gpt-neo-1.3B,
    gpt-neo-2.7B \citep{gptneo},
    gemma-2b \citep{gemmateam2024gemmaopenmodelsbased},
    phi-2 \citep{javaheripi2023phi},
    gpt-j-6B \citep{gpt-j},
    gemma-7b \citep{gemmateam2024gemmaopenmodelsbased},
    Llama-2-7b-chat-hf \citep{touvron2023llama2openfoundation},
    Mistral-7B-v0.1 \citep{jiang2023mistral7b},
    and
    Meta-Llama-3-8B \citep{dubey2024llama3herdmodels})
We frame the prompt as a multi-choice question-answering (MCQA) objective \citep{robinson2023leveraging}. 
The prompt is intentionally structured so that the LLM is intended to predict a single choice token (Such as `` A'', `` B'', etc.). \citet{robinson2023leveraging}  highlight the advantages of MCQA-based evaluation over cloze evaluation \citep{incontextfewshotlearners}(where the LLMs are expected to generate the entire answer in a cloze test), leading to a significant boost in various tasks, including commonsense-based tasks. 
App. \ref{app-sec:prompts}, Fig. \ref{fig:prompt_template_autoregressive} presents various prompt templates for autoregressive experiments, and 
App. \ref{app-sec:prompts} Fig. \ref{fig:prompt_template_autoregressive_qualitative} 
shows a few qualitative examples for the framed causal query templates.
    Table \ref{tab:LLM_evaluation_results} shows the success rate obtained for various LLMs. 
    The success rate corresponds to the percentage of queries where the LLM predicts the desired choice.
    We observe that reasoning causally about simple daily activities is challenging when a rigorous test is framed, validating the dependencies between the events. Overall, for the more common activities like \texttt{baking a cake} and \texttt{going grocery shopping}, the LLMs perform better when compared to activities like \texttt{boarding a bus} or \texttt{planting a tree}. We also experimented with another version of the dataset, where incorrect choice may correspond to temporally plausible but causally implausible events. The results drop significantly in this case; details and results are provided in the App. \ref{app-sec:temporal-results}.

\begin{table*}[t]
\caption{The table provides evaluation results of language models over the created causal triplets.}
\label{tab:LLM_evaluation_results}
\centering
\small
\renewcommand{\arraystretch}{1}
\setlength\tabcolsep{5pt}
{%
\begin{tabular}{clccccc}
\toprule
\textbf{Triplets}                                  & \textbf{Model Name} & \texttt{\textbf{cake}} & \texttt{\textbf{shopping}} & \texttt{\textbf{train}} & \texttt{\textbf{tree}} & \texttt{\textbf{bus}} \\ \midrule
\multirow{10}{*}{\rotatebox[origin=c]{90}{causal triplets}}  & gpt-neo-125M            & 50.71 & 50.01 & 49.99 & 50.13 & 50.15         \\
                                                       & gpt-neo-1.3B     &    44.77 & 45.69 & 42.52 & 45.67 & 42.89        \\
                                                       & gemma-2b        & 53.76 & 52.19 & 60.57 & 60.71 & 53.64        \\
                                                       & gpt-neo-2.7B  & 50.00 & 50.01 & 50.00 & 50.01 & 50.00        \\
                                                       & phi-2     & 85.14 & 83.65 & 77.29 & 82.24 & 71.74        \\
                                                       & gpt-j-6B     & 49.59 & 50.02 & 50.29 & 49.92 & 49.93         \\
                                                       & Llama-2-7b-chat-hf               & 77.92 & 72.41 & 73.48 & 72.40 & 68.21        \\
                                                       & Mistral-7B-v0.1 & 77.64 & 69.38 & 68.46 & 72.43 & 69.37        \\ 
                                                       & gemma-7b
                                                       & 81.47 & 82.26 & 77.24 & 80.78 & 70.29       \\ 
                                                       & Meta-Llama-3-8B
                                                       & 80.79 & 76.46 & 76.08 & 78.21 & 67.39        \\
\bottomrule
\end{tabular}}
\end{table*}

\noindent\textbf{Evaluation using Average Treatment Effect (ATE) ($\Delta$):} Computing the Average Treatment Effect ($\Delta$) helps establish the strength of causal links given a context (Eq. \ref{eq:average_treatment_effect}). In our setup, to estimate $P (y|do (x))$ (i.e., the causal estimand) from statistical estimands (obtained from observational distribution), we make certain reasonable assumptions about the underlying process that governs the relation among variables/events and then utilize the implications of these assumptions. For any activity taking place, the causal relationships between two events $E_1$ and $E_2$ may have a causal link along with a non-causal link through a set of confounders $z$. We define the confounder $z = \{t_i | t_i \in \mathcal{T}\}$, where $\mathcal{T}$ denotes all the trajectories (sequence of events) from the start of the activity till the event $E_1$. The temporal nature of events makes this assumption suitable since the occurrence of  $E_1$ and $E_2$ can be confounded by all the events preceding $E_1$. 
Note the possibility of unobserved confounders (events that are not explicitly mentioned but may be affecting the mentioned events) in our case is removed due to two reasons: 1) Keeping a closed system representation with a large number of diverse scripts (written by humans) helps cover the set of most generic and diverse events either implicitly or explicitly as a part of the activity, and 2) The causal reasoning goal is restricted to figuring out the causal effect between the events that are present explicitly. 
Assuming the unmentioned events have insignificant effects, we can establish that there will not be any unobserved confounders. This assumption makes the observed confounders satisfy the backdoor criterion \citep{26a5efafca6149a2a570d3ee5981ce1f} that make sufficient adjustment sets. By using the backdoor criterion (App. \ref{app-sec:backdoor}), the interventional distributions are estimated as follows: 
\begin{align}
P(E_2 | do  (E_1)) = \sum_{t_i \in \mathcal{T}} p_*(E_2| E_1 , z=t_i)p_*(z=t_i) \label{eq:backdoor}
\end{align}
\noindent Note that the true observational distribution, i.e. 
$p_*(E_2|E_1, z)$ and $p_*(z)$ both are unknown and have to be approximated  ($\hat{p}(E_2|E_1, z)$ and $\hat{p}(z)$). Further, we describe ways $\hat{p}$ can be estimated via multiple design mechanisms. Due to space limitation, we only describe the $\hat{p}$ estimation via language models below and move the statistical analysis using the original trajectories and observational graphs to the App. \ref{app-sec:ate-results}. 

\noindent\textbf{ATE using Language Models}
Since pre-trained LMs capture world knowledge \citep{devlin-etal-2019-bert, incontextfewshotlearners,li2023emergent, nanda-etal-2023-emergent,karvonen2024emergent}, these provide a suitable proxy for establishing relationships between these events. 
For our experiments, we consider a simple reasoning capability of language models, i.e., to reason about the temporal order of various events, i.e., given an event, what is the likelihood of the occurrence of another event? We further ask if this can be used to estimate the causal relationship between the events (a similar strategy is used by \citet{pmlr-v162-zhang22am-rock} for zero-shot causal estimation). 
It is worth noting that for these activities about daily activities, one way to find causes is to establish the temporal likelihood of the events. 
For each of the multiple language models, we frame the temporal prediction differently. 

\noindent \textbf{Encoder-only Models:} For BERT-based models trained for mask token prediction, we model the temporal prediction using the probability assigned to the mask tokens ``\textit{before}'' and ``\textit{after}'' \citep{pmlr-v162-zhang22am-rock}. 
Given two events $E_1$ and $E_2$, the temporal link is predicted using a prompt like $E_1$ <mask> $E_2$, and the scores corresponding to the before and after tokens are collected. App. \ref{app-sec:results}, Fig. \ref{fig:prompt_template_before_after}, the top row highlights the prompt template used for BERT-based models. For encoder-only experiments, we consider RoBERTa MNLI \citep{liu2019roberta}.

\noindent \textbf{Decoder-only Models:} For other language models that are autoregressive in nature, we modify the prompt to predict the temporal order as the last token. 
We again use the MCQA-based prompting style to frame the temporal order query by providing ``\textit{before}" and ``\textit{after}" as the options in the prompt.
App. \ref{app-sec:results}, Fig.  \ref{fig:prompt_template_before_after}, the bottom row highlights the prompt template used for Decoder-only Models.

\noindent \textbf{Interventions:} 
We utilize the SUTVA assumption in the proposed framework to devise an intervention over a trajectory in natural language form. App. \ref{app-sec:results}, Fig. \ref{fig:prompt_template_intervention} shows the style of intervention made by an event ($E_1$) taking place ($do(E_1)$) or not taking place $do(\neg E_1)$.

\noindent Given the above strategies, LMs can be used to evaluate $\hat{p} (E_2| \neg E_1 , z=t)$, by feeding the prompt that contains $E_1$, $E_2$ and the $z=t)$ and predict the temporal nature between $E_1$ and $E_2$, given a trajectory $z=t$. Further, applying the backdoor criterion for multiple trajectories $\mathcal{T}$, we obtain
\begin{align}
\begin{split}
    p_{\mathcal{M}} (E_{2}|do (E_1)) &= \frac{1}{|\mathcal{T}|}\sum_{t \in \mathcal{T}} \hat{p} (E_2| E_1 , z=t)\\
    p_{\mathcal{M}} (E_{2}|do (\neg E_1)) &= \frac{1}{|\mathcal{T}|}\sum_{t \in \mathcal{T}} \hat{p} (E_2| \neg E_1 , z=t)
\end{split}
\end{align}
which can further be used to estimate the causal strength between the events $E_1$ and $E_2$.
{
\begin{align*}
    \Delta_{\mathcal{M}} &= p_{\mathcal{M}} (E_{2}|do (E_1))-p_{\mathcal{M}} (E_{2}|do (\neg E_1))
\end{align*}
}

Using the multiple $\Delta$ estimates defined above, we estimate the causal strength between the events available for an activity. We follow the scheme presented in the App. \ref{app-sec:algorithms}, Algorithm \ref{alg:dataset_evaluation} to compute the performance in terms of success rates. 

\noindent\textbf{Temporal Scheme:} In this scheme, we validate if temporal ordering knowledge of LLMs could be directly used to estimate the causal estimand. We make use of templates shown in the App. Fig. \ref{fig:prompt_template_before_after}. The causal estimand is estimated via the difference in logit values when intervening over an event, i.e., does the predicted probability take into account the context of events not happening? Surprisingly, we found that temporal ordering does provide a suitable proxy for estimating causal strength between the events.  We further extend this approximation to incorporate the backdoor adjustments in the $\Delta$. 

\begin{table*}[t]
\caption{Accuracy over the causal triplets for various $\Delta$ estimates. The \textcolor{blue}{blue text} denotes the improvements by backdoor adjustments over the temporal scheme for multiple language models. \textbf{Bold text} represents the best-performing method for a particular activity.}
\label{tab:results}
\centering
\small
\renewcommand{\arraystretch}{1}
\setlength\tabcolsep{3pt}
\begin{tabular}{lclccccc}
\toprule
$\hat p$ \textbf{estimation} & \textbf{Scheme}      & \textbf{ATE} & \texttt{\textbf{Cake}} &  \texttt{\textbf{Shopping}} & \texttt{\textbf{Train}} & \texttt{\textbf{Tree}} & \texttt{\textbf{Bus}}  \\ \midrule
Original Trajectories                 &  & $\Delta_o$               &   28.20        & 34.30             & 31.10        & 30.10        & 30.40              \\ \midrule
\multirow{2}{*}{Observational Graphs} & - & $\Delta_n$               & 30.40        & 30.10             & 29.80          & 28.60         & 25.4            \\
                                      & - & $\Delta_t$               &  40.90         & 47.10             & 40.30          & 37.60         & 40.10               \\ \midrule
                                      \multirow{11}{*}{Language Models}  & \multirow{11}{*}{\rotatebox[origin=c]{90}{Temporal}} & RoBERTa MNLI & 46.80         & 54.00             & 45.50          & 52.70         & 43.00      \\

&  &     gpt-neo-125M         &  47.70 & 55.50 & 55.50 & 53.60 & 48.20          \\
                                      
&  &  gpt-neo-1.3B             &   47.40       &           45.40   &        53.30  &  43.40 &  52.90             \\
    &                                    &  gemma-2b            &    43.80  &    41.70      &      52.20   &     49.70    &   49.80     \\
                                      &  & gpt-neo-2.7B              & 50.10   &    48.90   &  52.40  &   47.60  &         53.70         \\
                                      & & phi-2             & 60.30  & 59.20   &56.90  &  70.30  & 49.40            \\
                                       &  & gpt-j-6B             & 49.50   &   46.40   &  56.00       &   62.70     &  56.00           \\
                                      &  & Llama-2-7b-chat-hf             &    38.90 & 42.10 &   51.00  &  40.70   & 47.80         \\
                                      
                                      & & Mistral-7B-v0.1                 &50.90    & 54.40     &   64.50   & 60.50    &  62.30             
                                      \\
                                      & & gemma-7b                 & 46.8 &54.00    & 45.50   & 52.70    &   43.00            
                                      \\
                                      & & Meta-Llama-3-8B                  & 58.20    & 54.10   & 55.6    &  55.00     &   64.00            
                                      \\
                                      \midrule
\multirow{11}{*}{Language Models}            & \multirow{11}{*}{\rotatebox[origin=c]{90}{Backdoor Adjustments}}& $\Delta_\mathcal{M}$ (RoBERTa MNLI)            & \textcolor{blue}{59.20} & \textcolor{blue}{54.40} & \textcolor{blue}{56.30} & \textcolor{blue}{57.50} & \textcolor{blue}{53.30} \\

 &  & $\Delta_\mathcal{M}$ (gpt-neo-125M)             &   \textcolor{blue}{59.20}     & {55.10}   &  {50.50}    & {52.10}    & 45.50          \\
                         &  & $\Delta_\mathcal{M}$ (gpt-neo-1.3B)             &   51.30       &  \textcolor{blue}{50.70}    & \textcolor{blue}{55.00}     & 43.90     & \textcolor{blue}{49.00}              \\
                                      &  & $\Delta_\mathcal{M}$ (gemma-2b)            & \textcolor{blue}{44.50}      &   \textcolor{blue}{45.30}& \textcolor{blue}{52.60}  & \textcolor{blue}{63.50}     &  43.90          \\
                                      &  & $\Delta_\mathcal{M}$ (gpt-neo-2.7B)              &   49.10 & \textcolor{blue}{51.30}  &  \textcolor{blue}{51.50}      &       \textcolor{blue}{54.00}   & \textcolor{blue}{51.40}    \\
                                      &  & $\Delta_\mathcal{M}$ (phi-2)             & 57.00   & \textcolor{blue}{66.00}   &   \textcolor{blue}{62.10}      &  57.10   &   45.80          \\
                                       &  & $\Delta_\mathcal{M}$ (gpt-j-6B)             & \textcolor{blue}{51.30}       & 45.60   & 50.50      &   49.10    & 46.00      \\
                                      
                                      &  & $\Delta_\mathcal{M}$ (Llama-2-7b-chat-hf)                 &  \textcolor{blue}{62.60} & \textcolor{blue}{64.60}    &    \textcolor{blue}{68.50}      &   \textcolor{blue}{70.50}     & \textcolor{blue}{63.80}            \\
                                      &  & $\Delta_\mathcal{M}$ (Mistral-7B-v0.1)        &  \textcolor{blue}{63.90} & \textcolor{blue}{71.40}    &    \textbf{\textcolor{blue}{73.70}}     &   \textcolor{blue}{61.30}     & \textbf{\textcolor{blue}{67.00}}             \\&  & $\Delta_\mathcal{M}$ 
                                      (gemma-7b) &  \textbf{\textcolor{blue}{72.80}} & \textbf{\textcolor{blue}{77.80}}    &    \textcolor{blue}{73.60}     &   \textbf{\textcolor{blue}{71.90}}    & \textcolor{blue}{62.40}          \\&  & $\Delta_\mathcal{M}$ (Meta-Llama-3-8B)                 &  \textcolor{blue}{66.00} & \textcolor{blue}{70.20}    &    \textcolor{blue}{68.40}     &  \textcolor{blue}{62.00}    & 63.40            \\
                                      \bottomrule
\end{tabular}
\end{table*}

\noindent\textbf{Backdoor Adjustments:} For the experiments with Language models, we apply the backdoor adjustment to estimate the causal estimand $\Delta_{\mathcal{M}}$. App. Fig. \ref{fig:prompt_template_intervention} shows the prompt template used to determine the relationship between the events. The prompt template takes a trajectory, $t_i$, that contains all the events till the event $E_1$ in sequential order of occurrence, further, an added prompt determines the intervention ($do(E_1)$ or $do(\neg E_1)$) and the causal estimand is estimated using the logit values associated with the predicted token. App. \ref{app-sec:algorithms}, Algorithm \ref{alg:causal_estimand_e1_e2} provides the designed scheme to compute unbiased causal estimands. We essentially flip the options and generate the scores associated with options `A' and `B' for increase and decrease, respectively (more details in the App. \ref{app-sec:backdoor}.) 

Table \ref{tab:results} shows a comparison between various design choices. We observe that when using LLMs for $\hat{p}$ estimation, the backdoor adjustment increases the performance over the temporal estimation scheme by a significant margin. The understanding of these activities is generic, and LLMs do provide a suitable set of sequences when prompted to generate a list of steps to complete the activity. For example, when prompted with \texttt{`Generate the sequential steps in a telegrammic style to perform the activity ``going grocery shopping"'}, almost all the models we tested provide a valid set of steps for the given activity. However, when prompted with causal queries, the lower performance signifies the lack of understanding of the underlying causal mechanism. 
The constructed dataset helps to rigorously validate the understanding of the activity through an enormous number of causal query triplets. The results show that although the LLMs can explain the activity in detail, including generating correct steps for performing tasks, causally reasoning about the set of events remains challenging.

\noindent\textbf{Human Study:} We conducted a small-scale human validation study over the created causal query dataset and asked 5 graduate students to answer 100 randomly sampled causal query triplets (20 per activity). We record an average performance of 92.20\%. (More details about the human study are provided in the App. \ref{app:human-study})

%% file: sections/relatedwork.tex
\section{Related Work}

Causal reasoning has been an active research area in the ML community \citep{Spirtes2000, PetersJanzingSchoelkopf17,causal_representation_learning_bernhard}. Some of the initial works highlight the causal nature of events present in text \citep{SCHANK1975237} as \textit{`causal chains'}. Multiple works have considered creating benchmarks/datasets that capture causal relationships between the events described in the text (see App. Table \ref{tab:causalbenchmarks}). More recently, with the rapid growth of LLMs on reasoning/understanding tasks, attention has shifted to validating these general-purpose models capturing causal reasoning \citep{jin2023largecorr2cause,2023causalparrots,willig2023probing,liu-etal-2023-magic,pmlr-v208-willig23a,Zhang2022OnTP,jin2024cladder}. 
App. \ref{app:data-comparison}
Table \ref{tab:causalbenchmarks} shows a broad overview of the existing causal Dataset/Benchmarks presented in the NLP community.
In this work, the primary focus is to bridge the gap between various lines of work that consider natural language to learn/validate/reason about causal relationships between events. 

%% file: sections/limitations.tex
\section{Limitations and Future Directions} \label{sec:limitations}
One of the primary limitations of our work is the limited set of activities. Though the frameworks support generating exhaustive/enormous causal queries, finding general commonsense reasoning activities/tasks that are well understood by humans remains challenging. Moreover, creating a causal graph for an activity increases as we move toward more long-term tasks. However, as a general test of causal intelligence, our framework provides a suitable platform to validate the reasoning capabilities more rigorously. In the future, it would be interesting to sample trajectories from the observational distribution $\mathcal{G}_o^{a}$ to create a training dataset and check if causal reasoning ability can be acquired by language modeling objectives (including other variants like presented in \citet{lampinen2023passive}). We leave this detailed analysis for future endeavors. The proposed algorithm for causal triplet generation generates the simplest variant of causal queries in the form of causal triplets (also referred to as Pairwise Causal Discovery (PCD) task by \citep{chen2024causal}). More complicated causal queries can be generated, such as considering cases with common confounders, long/short causal chain dependency, etc. Moreover, taking formal definitions.
(i.e., using the formal causal inference language) causal queries inspired from \citet{jin2023largecorr2cause, jin2024cladder} can be framed for a more rigorous analysis. Being at the initial state, we stick to the simple causal queries that provide two choices, and the task is to choose the more plausible cause. The creation of underlying causal graphs provides endless possibilities for creating varied versions of causal queries. In this work, we only consider an unconditional version of $\textit{d}$-separation. In the future, the same causal graphs could be used to define more datasets for covering other rungs of the \textit{`causal ladder'} \citep{bookOfWhy}.

%% file: sections/conclusion.tex
\section{Conclusion}

In this paper, we proposed the \titleShort\ (\titleFull) framework for generating causal queries that can be used to rigorously evaluate LLMs. We performed extensive experimentation with LLMs for the task of Causal Commonsense Reasoning. Results indicate that LLMs are still far from a complete understanding of daily commonsensical activities and fail to answer causal queries when analyzed in an exhaustive manner. We believe this framework will provide a good platform for future research in understanding the causal reasoning abilities of LLMs.

%% file: sections/appendix.tex
\section*{Appendix}

\appendix


\hypersetup{linkcolor=blue}


\startcontents[appendix] 
\section*{Table of Contents} 
\printcontents[appendix]{section}{0}{\setcounter{tocdepth}{4}} 

\startlist[appendix]{lot} 
\section*{List of Tables} 
\printlist[appendix]{lot}{}{\setcounter{tocdepth}{2}} 

\startlist[appendix]{lof} 
\section*{List of Figures} 
\printlist[appendix]{lof}{}{\setcounter{tocdepth}{2}}

\newpage


\section{\titleShort\ Framework Details}

The \titleShort\ framework consists of Observational Distributions represented in the form of DAGs ($\mathcal{G}_o$) along with the corresponding causal graphs ($\mathcal{G}_c$) governing the dependency of occurrence between the events. Table \ref{tab:graph_details_and_results} highlights the total number of causal queries that can be created using the framework. Table \ref{tab:dataset:eg} shows a qualitative comparison with the COPA dataset \cite{gordon-etal-2012-semeval_copa_dataset} and the triplet samples coming from the \titleShort\ framework.

\subsection{Adherence to SUTVA} \label{app-sec:sutva}

In causal literature, a fundamentally acknowledged \emph{Stable Unit Treatment Value Assumption} (SUTVA) \citep{Cox1958-COXPOE, Rubi:80}) requires that for each unit (e.g., sequence of events), there is only one version of the non-treatment, i.e., for an event in the sequence, there lie only two versions occurring and not occurring. SUTVA plays a vital role in causal inference by ensuring that each unit's treatment assignment has a consistent impact, facilitating the accurate estimation of treatment effects. Although, in the past, researchers have created various datasets that capture the causal relationship between real-world events \citep{gordon-etal-2012-semeval_copa_dataset, du-etal-2022-e}, the problem of achieving the SUTVA assumption has remained challenging. For example, given events (taken from the COPA dataset \citep{gordon-etal-2012-semeval_copa_dataset}) $E_1$: \texttt{“The teacher assigned homework to students”} and $E_2$: \texttt{“The students groaned,”} it becomes challenging to define $\neg E_1$ since there are enormous possibilities that may have occurred at the same time (in place of $E_1$) that negates $E_1$, making it difficult to define an event of not having done something. Recent work by \citet{pmlr-v162-zhang22am-rock}, proposes to use multiple alterations of events for capturing $\neg E_1$, violating the SUTVA assumption.  In this work, we highlight that if we define a closed system, capturing a commonsense activity, it facilitates adherence to SUTVA assumptions as closely as possible.
For example, in the activity of \texttt{``going via an airplane,”} one would have either \texttt{“checked-in the luggage”} ($E_1$) or \texttt{``skipped checking-in luggage"} due to smaller bags ($\neg E_1$). Moreover, developing a causal setup with observations has always been a challenging problem in the wild and often requires few assumptions, as the strong causal link can only be established in an ideal world where randomized controlled trials (RCTs) are feasible. 
In our framework, adhering to SUTVA comes naturally where, in a trajectory, an occurrence of an event can be intervened to obtain an alternate trajectory, reaching an ideal setup facilitating causal reasoning in daily commonsensical activities. 

\subsection{Comparison with previous Causal Reasoning Datasets/Benchmarks} \label{app:data-comparison}
Table \ref{tab:causalbenchmarks} shows a broad overview of the existing causal Dataset/Benchmarks presented in the NLP community. We find that most of the existing set of work relies on real-world events to reason about causality in NLP, where human annotators are asked to reason causally between the nature of events. However, most of these datasets/benchmarks try to establish a connection using a simple question prompt, which may not be enough to construct the underlying causal graph. Moreover, most of the real-world grounding-based methods remain open-ended due to the events taking place in the wild, making it difficult to consider constructing a causal graph where multiple variables play a role. More recently, with increased research attention on the causal reasoning abilities of LLMs, researchers have tried framing causal queries based on a causal inference engine, requiring the underlying causal graphs. However, when constructing causal queries from prompting LLMs, natural language is used to verbalize the causal concepts in the form of symbolic variables that may not have a real-world grounding.\footnote{https://huggingface.co/datasets/causalnlp/corr2cause} Moreover, the created causal queries are difficult for a human with little or no knowledge of causal inference concepts.\footnote{https://huggingface.co/datasets/causalnlp/CLadder} 
Table \ref{tab:causalbenchmarks} shows a comparison of all these features in detail, where \titleShort\ satisfies all the features. 

We realize this is a first-of-its-kind framework built over real-world events and contains the underlying causal graph. Having both the Observational Distribution (representing the enormous event sequences present in daily activity) and the manually created underlying causal graph helps facilitate an in-depth analysis of the causal reasoning abilities of LLMs. Moreover, the same framework can further be extended in various ways: 1) Extending the number of activities: In the current version of the framework, we only consider 5 daily activities to provide an in-depth analysis. In the future, those can be extended to incorporate more such activities. 2) Extending the scope of activities: The tasks used in the activities are generic and capture commonsense; for validating domain-specific causal reasoning abilities, the framework could be extended to domain activities, for example, cooking a specific recipe where adding different ingredients causes a variation in taste.
3) Extending the type of causal queries: while constructing the causal queries, we considered the simplest task of finding the more plausible cause/effect given two options as events, keeping only unconditional $d-$separation as the primary condition. The framework can directly be extended, keeping causal queries inspired from \citet{jin2023largecorr2cause, jin2024cladder}.

On the analysis front, we realize that the possibilities of an in-depth analysis increase by a significant margin. In this work, we shed light on a few mechanisms for validating causal reasoning abilities via zero-shot CCR (compared to previous works that rely on training and further testing on similar datasets). We specifically focused on open-weight models for better applicability in the future and proposed a few mechanisms for estimating the causal relationships between the events. This opens up several possible avenues for an in-depth analysis of LLMs. 

\subsection{Observational Graphs} \label{app-sec:observation-graph}

\begin{figure}[t]
\centering
  \includegraphics[width=0.9\linewidth]{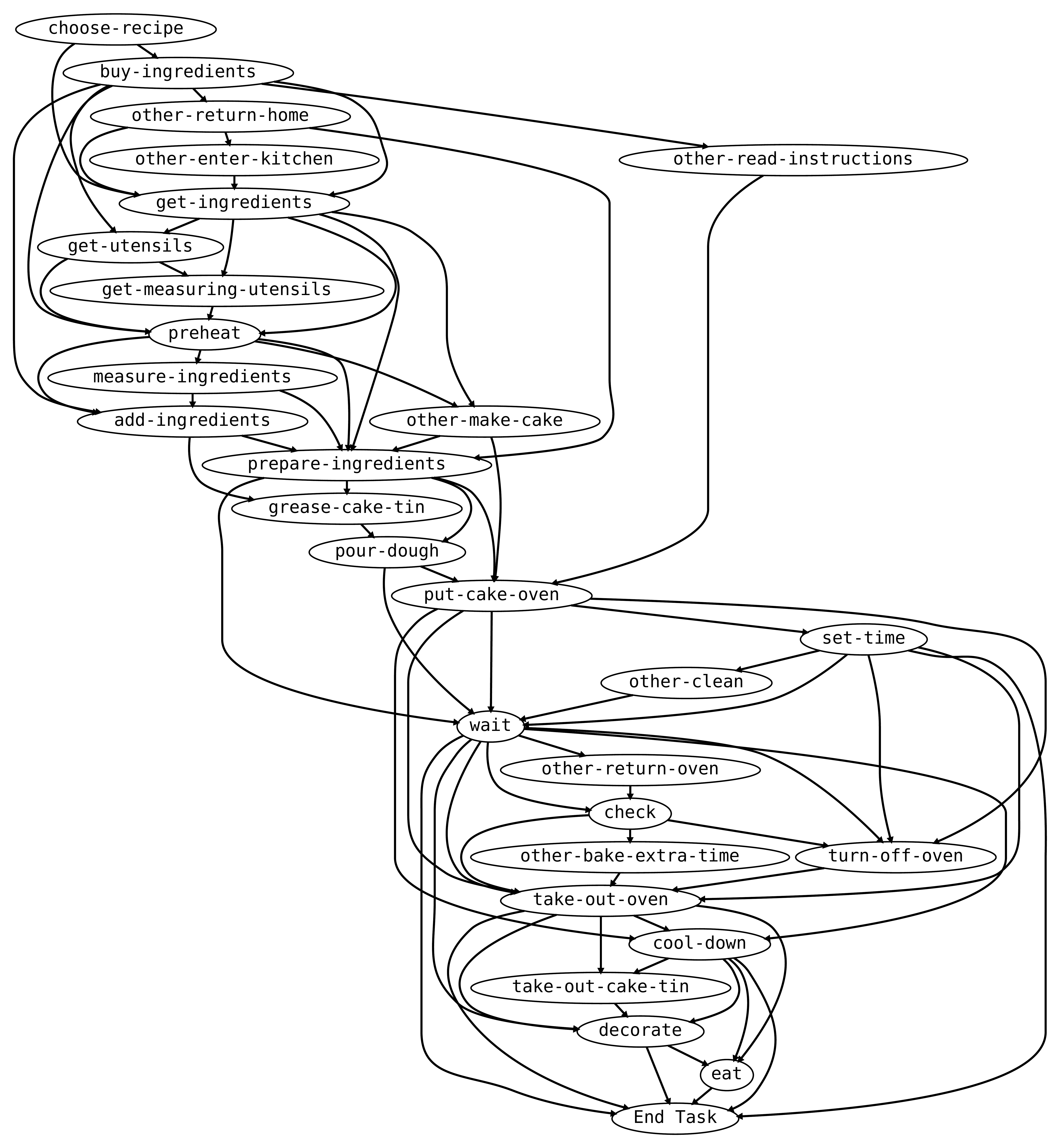}
  \caption[The \textit{``observational graph"} for the activity \texttt{Baking a Cake}.]{{The figure shows the \textit{``observational graph"} for the activity \texttt{Baking a Cake}. }}
  \label{fig:baking_a_cake}
\end{figure}

\begin{figure}[t]
\centering
  \includegraphics[width=0.9\linewidth]{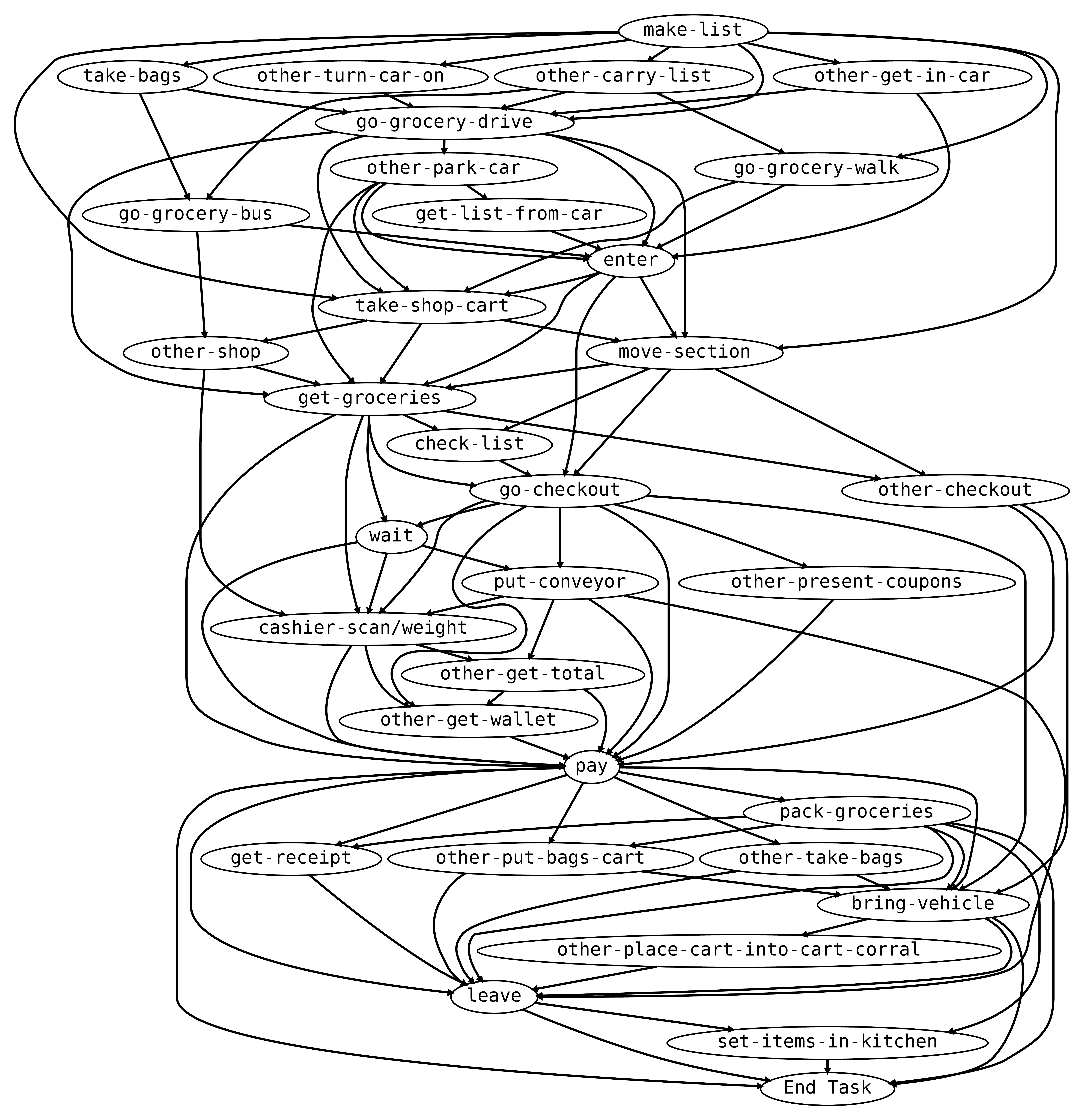}
  \caption[The \textit{``observational graph"} for the activity \texttt{Going Grocery Shopping}.]{{The figure shows the \textit{``observational graph"} for the activity \texttt{Going Grocery Shopping}. }}
  \label{fig:going_grocery_shopping}
\end{figure}

\begin{figure}[t]
\centering
  \includegraphics[width=0.8\linewidth]{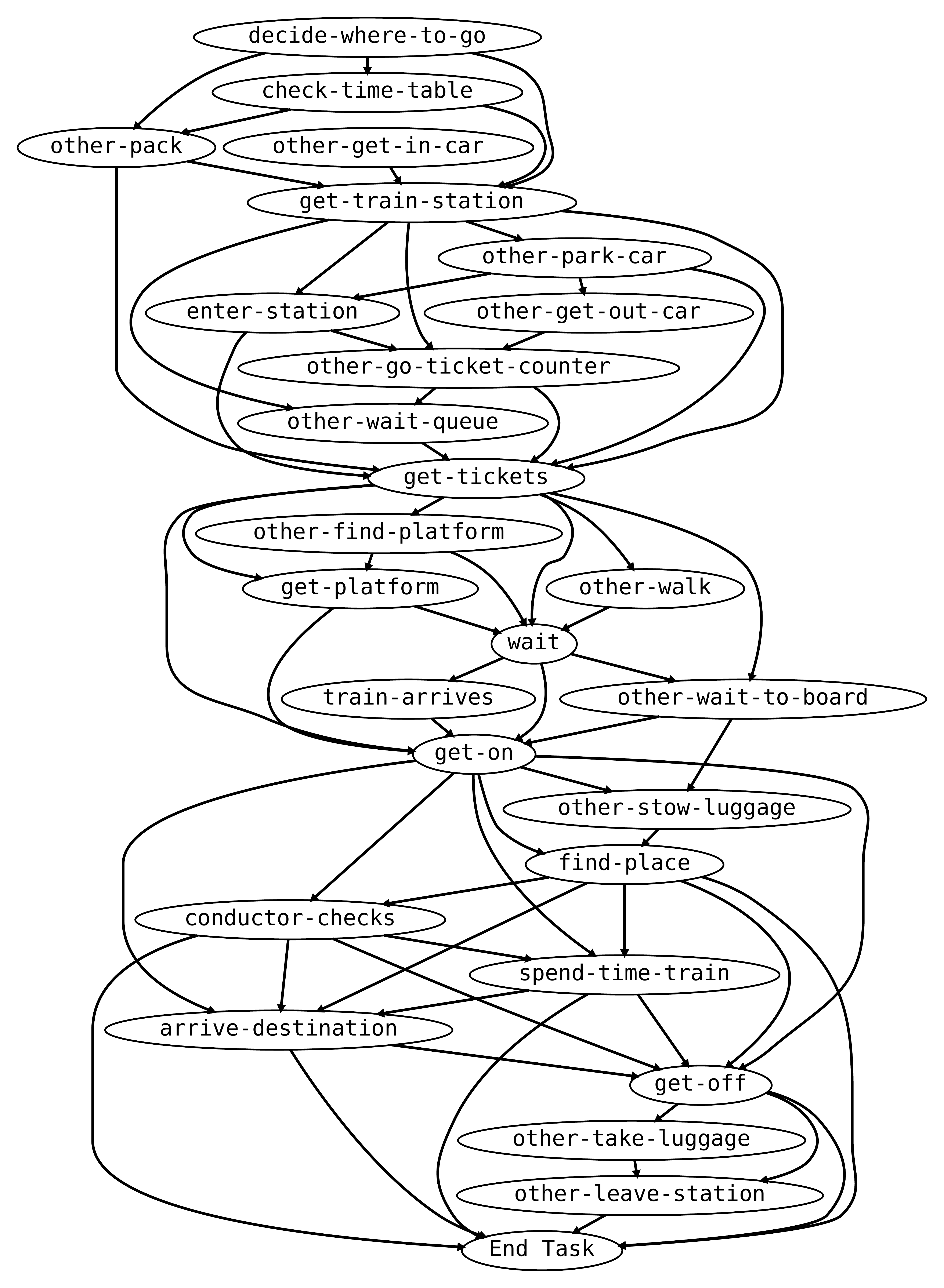}
  \caption[The \textit{``observational graph"} for the activity \texttt{Going on a Train}.]{{The figure shows the \textit{``observational graph"} for the activity \texttt{Going on a Train}. }}
  \label{fig:train_compact_graph_train}
\end{figure}

\begin{figure}[t]
\centering
  \includegraphics[width=0.98\linewidth]{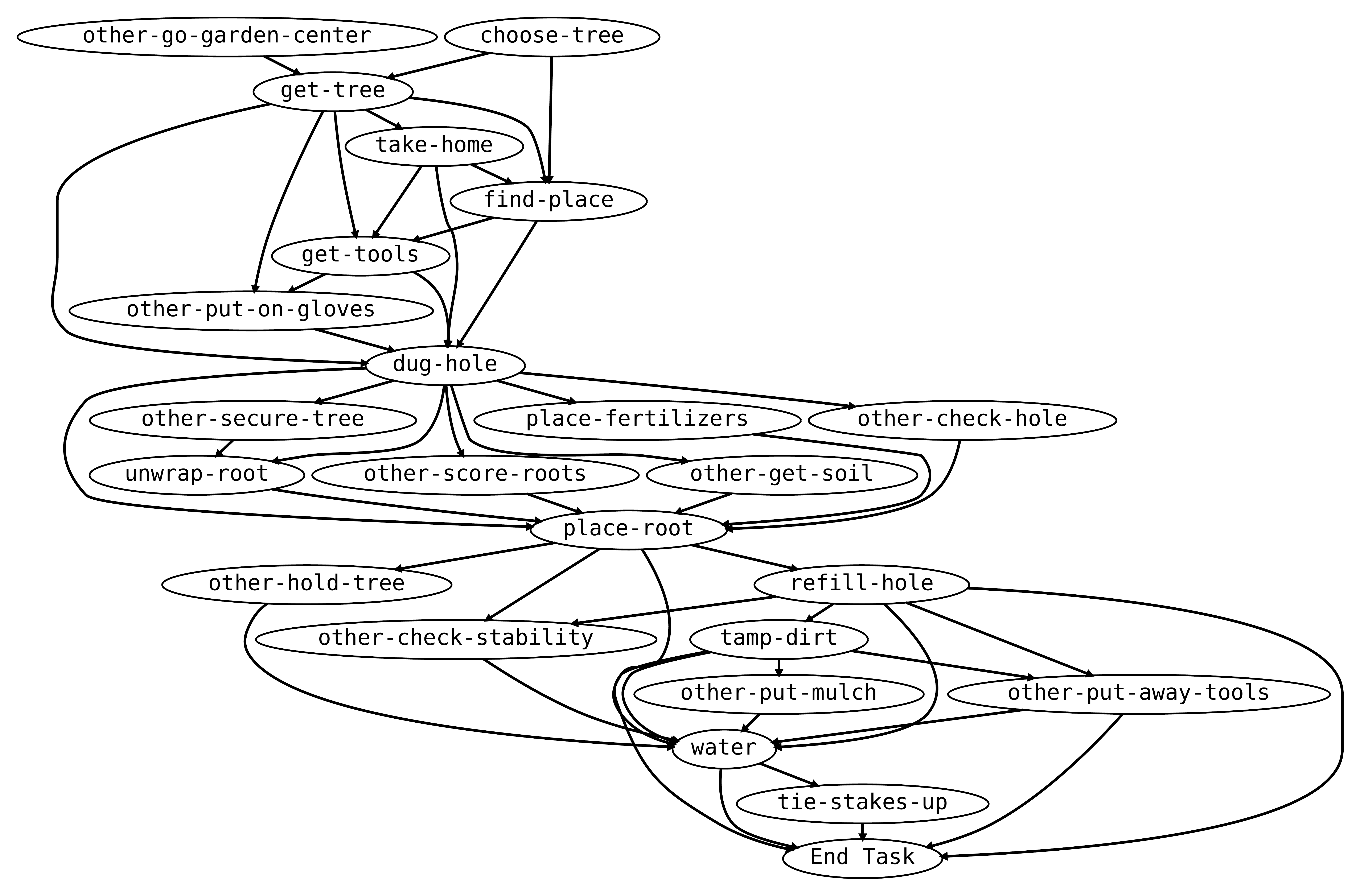}
  \caption[The \textit{``observational graph"} for the activity \texttt{Planting a Tree}.]{{The figure shows the \textit{``observational graph"} for the activity \texttt{Planting a Tree}. }}
  \label{fig:planting_a_tree}
\end{figure}

\begin{figure}[t]
\centering
  \includegraphics[width=0.75\linewidth]{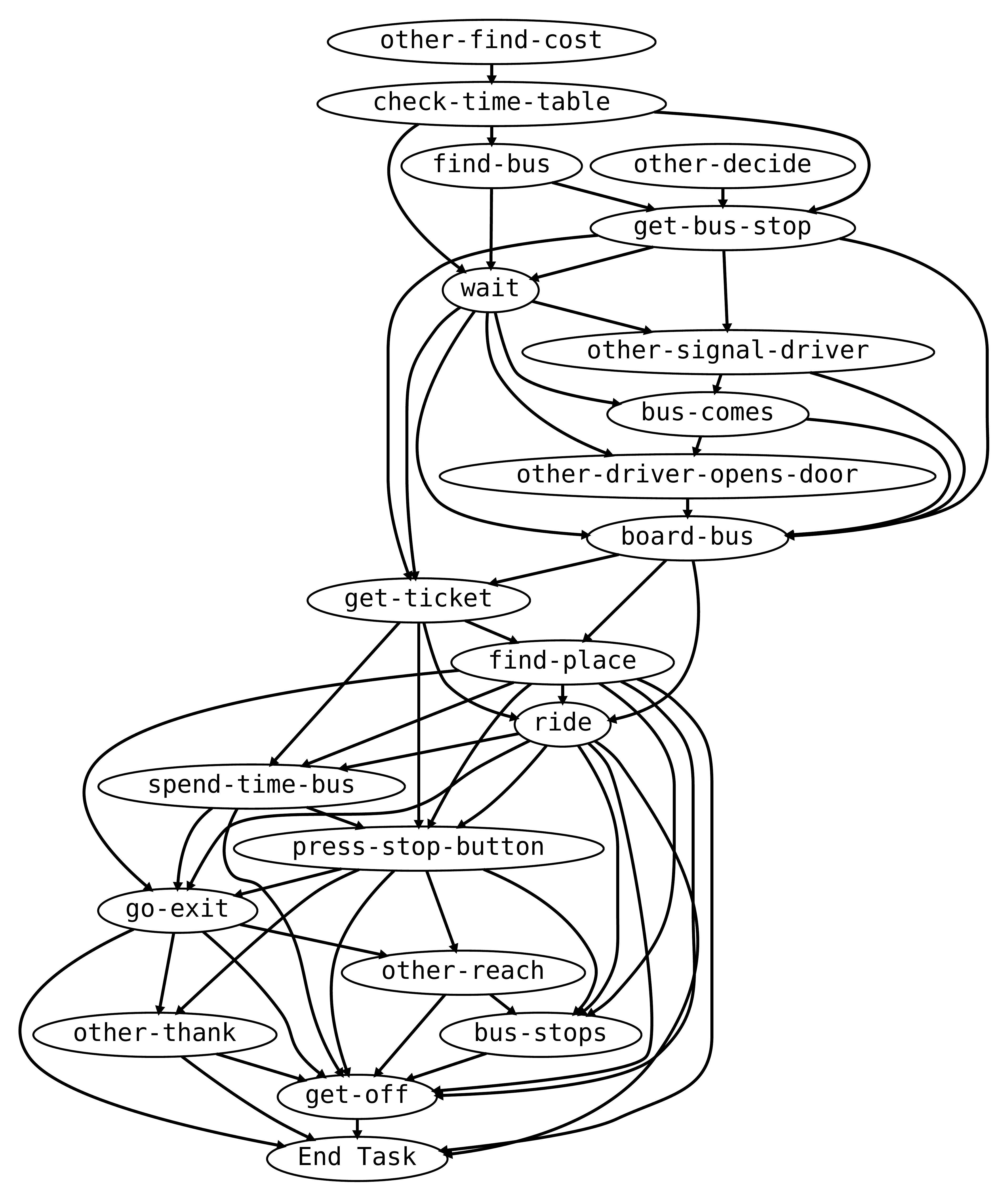}
  \caption[The \textit{``observational graph"} for the activity \texttt{Riding on a Bus}.]{{The figure shows the \textit{``observational graph"} for the activity \texttt{Riding on a Bus}. }}
  \label{fig:riding_on_a_bus}
\end{figure}

Fig. \ref{fig:baking_a_cake}, Fig. \ref{fig:going_grocery_shopping}, Fig. \ref{fig:train_compact_graph_train}, Fig. \ref{fig:planting_a_tree}, and Fig. \ref{fig:riding_on_a_bus} shows the \textit{``observational graphs"} for the activity \texttt{Baking a Cake}, \texttt{Going Grocery Shopping}, \texttt{Going on a Train}, \texttt{Planting a Tree}, and \texttt{Riding on a Bus} respectively.

\begin{table*}[!t]
	\centering
	\caption[Comparison with existing Causal Benchmarks/Datasets in NLP]{
 \small
		\textbf{Comparison of causal experimental settings used in prior LLM evaluation benchmarks.}
		The real-world grounding plays a crucial role in evaluating LLMs, which is not present in the symbolic benchmarks.
	}
	\label{tab:unfair_comparison}
	\resizebox{1.\textwidth}{!}{%
		\begin{tabular}{lccccr}
			\toprule
			\textbf{Datasets/Benchmarks}                         & \textbf{Real-World } & \textbf{Causal Graph} & \textbf{Symbolic} & \textbf{Exhaustive}  & \textbf{\# Samples}      \\ \midrule
   SemEval2021 Task8 \citep{hendrickx-etal-2010-semeval} & \cmark & \xmark & \xmark & \xmark & 1331 \\
   EventCausality \citep{do-etal-2011-minimally} & \cmark & \xmark & \xmark & \xmark & 414 \\
   COPA~ \citep{gordon-etal-2012-semeval_copa_dataset} & \cmark & \xmark & \xmark & \xmark & 1000 \\
   Causal-TimeBank \citep{mirza-etal-2014-annotating} & \cmark & \xmark & \xmark & \xmark & - \\
   CaTeRS \citep{mostafazadeh-etal-2016-caters} & \cmark & \xmark & \xmark & \xmark &  320 stories (1.6K sent) \\
   BECauSE \citep{dunietz-etal-2017-corpus} & \cmark & \xmark & \xmark & \xmark & - \\
   Event2Mind \citep{rashkin-etal-2018-event2mind} & \cmark & \xmark & \xmark & \xmark & 25K event phrases \\
   
   ATOMIC \citep{sap2019atomic} & \cmark & \xmark & \xmark & \xmark & 877K \\
   
   SocialIQA \citep{Sap2019SocialIQA} & \cmark & \xmark & \xmark & \xmark & 37K \\
   
   TimeTravel \citep{qin-etal-2019-counterfactual} & \cmark & \xmark & \xmark & \xmark & 81.4K \\
   
   Abductive (ART) \citep{bhagavatula2020abductive}  & \cmark & \xmark & \xmark & \xmark & 20K
narr, 200K expl  \\
   Com2Sense \citep{singh-etal-2021-com2sense} & \cmark & \xmark & \xmark & \xmark &  4k sentence pairs. \\
   TellMeWhy \citep{lal-etal-2021-tellmewhy} & \cmark & \xmark & \xmark & \xmark & 30K questions \\
   CRASS \citep{frohberg-binder-2022-crass} & \cmark & \xmark & \xmark & \xmark & 274 PCT \\
   e-CARE \citep{du-etal-2022-e} & \cmark & \xmark & \xmark & \xmark &  20K CR questions \\ 
   CausalQA \citep{bondarenko-etal-2022-causalqa} & \cmark & \xmark & \xmark & \xmark &  1.1 Million \\
   COLA \citep{wang-etal-2023-cola} & \cmark & \xmark & \xmark & \xmark &  1,360 event pairs \\
   CRAB \citep{romanou-etal-2023-crab} & \cmark & \xmark & \xmark & \xmark &  2.7K pairs \\
   
   \midrule
   CausalJudgement~ \citep{srivastava2023beyondcausaljudgement} & \cmark & \cmark & \xmark & \xmark &  - \\
Corr2Cause~ \citep{jin2023largecorr2cause} & \xmark & \cmark & \cmark & \cmark &  200K \\
CausalParrots~ \citep{2023causalparrots} & \cmark & \cmark & \cmark & \xmark & - \\
CLadder~ \citep{jin2024cladder} & \xmark & \cmark & \cmark & \xmark & 10K\\ \midrule

\textbf{COLD (ours)}~  & \cmark   & \cmark & \cmark & \cmark & $\sim$9.52 Million \\

			\bottomrule
		\end{tabular}
	}
	\label{tab:causalbenchmarks}
\end{table*}


\section{Algorithms in the \titleShort\ Framework} \label{app-sec:algorithms}

In this section, we provide insights into the Algorithms used in the \titleShort\ framework. We start with Algorithm \ref{alg:dataset_creation}, which creates causal query triplets given the observational graphs $\mathcal{G}_o$ and along with the Causal Graphs $\mathcal{G}_c$. 

\noindent\textbf{Remark:} Temporal precedence is generally assumed essential for defining causation, and it is one of the most important clues that is used to distinguish causal from other types of associations \citep{mill1898system, hill1965environment, pearl1995theory}. For our framework, we also consider the topologically sorted order obtained from the observational graphs and use the temporal order to define the causal query triplets, i.e., the cause events will always precede the effect events.

\noindent \textbf{Creating Causal Query Triplets:} The Algorithm \ref{alg:dataset_creation} is designed to sample all the possible causal query triplets to construct a dataset for validating causal reasoning ability over an activity. Provided the observational graphs $\mathcal{G}_o$ and the Causal Graphs $\mathcal{G}_c$ for an activity, we first sample all the possible node triplets in the graph. Later, we iterate over the set of triplets and check if one of the nodes in the triplet $(n_i, n_j, n_k)$ is {$\textit{d}$-separated}. Further, the {$\textit{d}$-separated} node becomes the wrong choice. The remaining two events become premise and correct choices, depending on their temporal order. For example, if $n_i$ is the node that is $\textit{d}$-separated from $n_j$ and $n_k$, we check if $n_j$ and $n_k$ have a causal link between them in $\mathcal{G}_o$. If $n_j$ and $n_k$ are found to have a causal link, we create two triplets using the temporal ordering between $n_j$ and $n_k$. The temporal link $n_j \rightarrow n_k$ leads to an `effect' query where $n_j$ becomes `premise' and $n_k$ becomes `correct choice,' and a `cause' query where $n_j$ becomes `correct choice' and $n_k$ becomes the `premise.' Note in Algorithm \ref{alg:dataset_creation} [Store tuple], we only show one such instance for brevity; the implementation will consist of another mirror instance (i.e., for every [Store tuple] both `cause' and `effect' question triplets are stored to the dataset).

The understanding of these activities is generic, and LLMs do provide a suitable set of sequences when prompted to generate a list of steps to complete the activity. The constructed dataset helps to rigorously validate the understanding of the activity through an enormous number of causal query triplets. The results show that although the LLMs can explain the activity in detail, including generating correct steps for performing tasks, causally reasoning about the set of events remains challenging.

\textbf{Human validation:} \label{app:human-study}
To get a rough estimate of the human performance on the created causal reasoning queries, we also perform a small-scale human study, where the annotators are given a set of randomly chosen 100 causal queries. 
The human subjects were graduate students of computer science who were given a brief tutorial about counterfactual reasoning.
Table \ref{tab:human_performance} shows the obtained results.
We would like to mention that validating human performance is challenging due to the nature of the causal reasoning task. The nature of counterfactual reasoning requires the human/algorithm to assume a similar alternate world/universe with only a particular happening or not happening to approximate the causal strength between the events. These imaginations can be expressed in statements as highlighted by \cite{bookOfWhy, pearl2016causal}, containing an “if” statement in which the “if” portion is untrue or unrealized (aka counterfactual). The “if” portion of a counterfactual is called the hypothetical condition, or more often, the antecedent, making it challenging (cognitively heavy) to conduct a human evaluation.
Please note that the study is performed using only a small sample of 100 causal query triplets out of thousands of queries, and the presented results only provide a rough estimate that may not generalize for a larger number of queries. Hence, a comparison of human study results and LLMs is not fair, and the presented human performance estimates may not be true representative of the entire population. Interactions with human subjects also revealed that they tend to confuse temporality and causality (similar findings were reported by \citet{do-etal-2011-minimally}).

\begin{table*}[!t]
	\centering
	\caption[Human performance]{
		\textbf{Human validation done for a small sample of 100 causal query triplets.}
		Overall we find that humans do perform well in causal reasoning about these daily activities.
	}
	\label{tab:human_performance}

\centering
\begin{tabular}{ccccccc}
\toprule 
Human Annotators               & \texttt{\textbf{cake}} & \texttt{\textbf{shopping}} & \texttt{\textbf{train}} & \texttt{\textbf{tree}} & \texttt{\textbf{bus}}   & Average \\ \midrule
Subject 1 & 95 & 95     & 90    & 100    & 90  & 94    \\
Subject 2 & 100    & 100        & 90    & 95 & 90  & 95    \\
Subject 3 & 100    & 100        & 85   & 85 & 70  & 88    \\
Subject 4 & 100    & 100        & 95   & 90  & 85 & 94    \\
Subject 5 & 100    & 85     & 95   & 90  & 80  & 90     \\
\midrule
Average & 99.00	& 96.00 &	91.00 &	92.00 &	83.00 &	\textbf{92.20} \\ \bottomrule
\end{tabular}
\end{table*}

\noindent \textbf{Evaluating over Causal Query Triplets: } Algorithm \ref{alg:dataset_evaluation} makes use of the causal estimands $\Delta$ to compare the causal strength between the premise event and the choice events. We consider the causal estimand computed between the premise and the available set of choices and predict the label corresponding to the high $\Delta$ values. 
For a given causal query from the created causal query triplet dataset $\mathcal{D}$, where each data entry $\mathcal{D}_i$ corresponds to $(p, c_1, c_2, q, l)$, i.e., premise event, choice 1, choice 2, question and the label respectively. As the task is to predict the more plausible cause/effect of the given premise event, we create two event pairs, $(p, c_1)$ and $(p, c_2)$, and compute the causal estimand $\Delta$ for both the pairs using the temporal or the backdoor scheme (described below in Algorithm \ref{alg:causal_estimand_e1_e2}). Note that the order of events given to $\Delta_{\mathcal{M}}$ is in $E_1$ and $E_2$ format, i.e. $\Delta_{\mathcal{M}}(E_1, E_2)$. Using the temporal precedence (highlighted as remark above), the cause event will always precede the effect event temporally. Hence, for a causal query with the question as `cause', the causal estimand is estimated as $\Delta_{\mathcal{M}}(c_1^i, p^i)$, $\Delta_{\mathcal{M}}(c_2^i, p^i)$ and $\Delta_{\mathcal{M}}(p^i, c_1^i)$, $\Delta_{\mathcal{M}}(p^i, c_2^i)$ when the question is `effect.' Further, based on the estimated $\Delta_{\mathcal{M}}$ scores, the more plausible cause/effect is predicted.

\noindent \textbf{Computing $\Delta_{\mathcal{M}}$:}
Algorithm \ref{alg:causal_estimand_e1_e2} depicts the process of computing an unbiased estimate for the causal estimand. 
The causal strength is computed between two events $E_1$ and $E_2$ where $E_1$ is assumed to be preceding $E_2$ temporally. To make an unbiased estimate based on the provided options, we consider normalizing the obtained probability scores by flipping the options and providing the same query prompt to the Language Model.
$$
\displaystyle f_{\mathcal{M}}(E_1,E_2,\phi)\leftarrow\frac{s_{\mathcal{M}}(E_1,E_2,\phi)+s_{\mathcal{M}}(E_1,E_2,\phi_f)}{s_{\mathcal{M}}(E_1,E_2,\phi)+s_{\mathcal{M}}(E_1,E_2,\phi_f)+\tilde{s}_{\mathcal{M}}(E_1,E_2,\phi)+\tilde{s}_{\mathcal{M}}(E_1,E_2,\phi_f)},
$$
where $\phi$ denotes the prompt template as shown in Figure \ref{fig:prompt_template_intervention} (top) and $\phi_f$ denotes the same prompt with flipped options, Figure \ref{fig:prompt_template_intervention} (bottom). The overall equation helps normalize the prediction probabilities of the `Increase' option by using the probabilities of the `Decrease' option. Finally, these normalized scores are computed for multiple trajectories $t_i$ in the backdoor adjustment scheme to compute the causal estimands $p_\mathcal{M}(E2 \mid do(E_1))$ and $p_\mathcal{M}(E2 \mid do(\neg E_1))$ that help estimate the causal strength $\Delta_{\mathcal{M}}$ between the events $E_1$ and $E_2$.

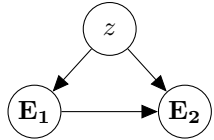
\begin{figure*}[h]
    \centering
    \begin{tikzpicture}[scale=1, transform shape]
        \node[latent] (e1) at (0,0) {$\mathbf{E_1}$};
        \node[latent] (e2) at (2,0) {$\mathbf{E_2}$};
        \node[latent] (z) at (1,1.1) {$z$};
        
        \draw[->] (z) -- (e1);
        \draw[->] (z) -- (e2);
        \draw[->] (e1) -- (e2);
    \end{tikzpicture}
\caption{ Causal Graphical Model of Events. $E_1$ temporally precedes $E_2$, and $z$ is trajectory variable, which assumes a values $t$ where $t \in$ All trajectories from start to $E_1$}
    \label{fig:causal_model_backdoor}
\end{figure*}


\begin{algorithm}[t]
\small
    \caption{Creating Causal Query Triplets}
    \label{alg:dataset_creation}
    \begin{algorithmic}
        \STATE $\mathcal{G}_c$:~ Causal Graphical Model; 
            $\mathcal{G}_o$:~ Observational Graph;
        \STATE $\mathbf{d_s(G,x,y)}$:~True iff $\mathbf{(x,y)}$ are \textit{d-separated} unconditionally in any DAG  $\mathbf{G}$;  
        \STATE $\mathbf{d_c(G,x,y)}$:~True iff $\mathbf{(x,y)}$ are \textit{d-connected} unconditionally in any DAG $\mathbf{G}$;
        
       \STATE $\mathbf{genSamples(G)}$:~ Generates all possible node triplets $(n_i, n_j, n_k)$ from DAG $\mathbf{G}$ such that $i\leq j \leq k$, where $(i, j, k)$ are the respective indices of nodes in a topologically sorted list of nodes;
       \STATE  $\mathbf{A(G,x,y)}$:~ True iff $\mathbf{x}$ is an ancestor of $\mathbf{y}$ in DAG $\mathbf{G}$.
        \STATE $\bm{\oplus}$:~ Exclusive OR operator;
        \STATE $p$: premise; $c_1$: choice 1; $c_2$: choice 2; $q$: question; $l$: answer (label)
        \STATE $(p, c_1, c_2, q, l) \in  \mathcal{D}$ 
        \STATE $\mathcal{D} = [\quad]$\hfill [Empty Dataset]

        \STATE $\mathcal{S} \sim \mathbf{genSamples(\mathcal{G}_o)}$\hfill [Generate Samples]

        \STATE \textbf{for} $(n_i , n_j , n_k )$ in $\mathcal{S}$
         \STATE~~~ \textbf{if} $\mathbf{d_s}$$(\mathcal{G}_c,n_i,n_j)$ \textbf{then}\\
         \STATE~~~~~~~~~ \textbf{if} $\mathbf{d_c}$ $(\mathcal{G}_c,n_i,n_k)$ $\bm{\oplus} $ $\mathbf{d_c}$ $(\mathcal{G}_c,n_j,n_k)$\\
        \STATE~~~~~~~~~~~~~~~$q=\text{`cause'}$\\
        \STATE~~~~~~~~~~~~~~~$l=\arg \max [\mathbf{d_c} 
        (\mathcal{G}_c,n_i,n_k) , 
        \mathbf{d_c}$$(\mathcal{G}_c,n_j,n_k)$]\\
        \STATE~~~~~~~~~~~~~~ $p,c_1,c_2 \leftarrow n_k, n_i,n_j$
        \STATE ~~~~~~~~~~~~~~~$\textbf{if}\quad\mathbf{A}(\mathcal{G}_c,c_ {l},p)\hspace{1mm}\textbf{then\quad}\mathtt{APPEND}(\mathcal{D},[(p , c_1, c_2 ,q,l ))$\hfill [Store tuple] \\
        \STATE~~~~\textbf{else if} $\mathbf{d_s}$$(\mathcal{G}_c,n_j,n_k)$ \textbf{then}\\
        \STATE~~~~~~~~~ \textbf{if} $\mathbf{d_c}$$(\mathcal{G}_c,n_i,n_j)$ $\bm{\oplus}$ $\mathbf{d_c}$$(\mathcal{G}_c,n_i,n_k)$\\
        \STATE~~~~~~~~~~~~~~~$q=\text{`effect'}$\\
        \STATE~~~~~~~~~~~~~~~$l=\arg \max [\mathbf{d_c} 
        (\mathcal{G}_c,n_i,n_k) , 
        \mathbf{d_c}$$(\mathcal{G}_c,n_i,n_j)$]\\
        \STATE~~~~~~~~~~~~~~ $p,c_1,c_2 \leftarrow n_i,n_j,n_k$
        \STATE ~~~~~~~~~~~~~~~$\textbf{if}\quad\mathbf{A}(\mathcal{G}_c,p,c_l)\hspace{1mm}\textbf{then\quad}\mathtt{APPEND}(\mathcal{D},(p , c_1, c_2 ,q,l ))$\hfill [Store tuple] \\
        \STATE \textbf{return} $\mathcal{D}$
    \end{algorithmic}
\end{algorithm}

\begin{algorithm}[t]
\small
\caption{Evaluating Causal Query Triplets}\label{alg:dataset_evaluation}
\begin{algorithmic}
    \STATE\hspace{-4mm}$\mathcal{T}_n$: $n$ unique trajectories from Start node (start of the activity) to node $E_1$
    \STATE $\displaystyle p_{\mathcal{M}} (E_2 | do (E_1)) = \frac{1}{|\mathcal{T}_n|}\sum_{t \in \mathcal{T}_n} \hat{p} (E_2 | E_1 , z = t)$
    \STATE $\displaystyle  p_{\mathcal{M}} (E_2 | do (\neg E_1)) = \frac{1}{|\mathcal{T}_n|}\sum_{t \in \mathcal{T}_n} \hat{p} (E_2 | \neg E_1 , z = t)$
    \STATE $\Delta_{\mathcal{M}}$: Returns \textbf{Average treatment effect} ($p_{\mathcal{M}} (E_2 | do (E_1)) - p_{\mathcal{M}} (E_2 | do (\neg E_1))$) to determine the causal effect of event $E_1$ on event $E_2$
    \STATE $p$: premise; $c_1$: choice 1; $c_2$: choice 2; $q$: question; $l$: label
    \STATE $\mathcal{D}$: Set of all causal queries
    \STATE $\mathcal{D}_i$: Causal query $(p, c_1, c_2, q, l)$, where $(p, c_1, c_2, q, l) \in \mathcal{D}$

    \FOR{$\mathcal{D}_i$ in $\mathcal{D}$}
        \IF{$q^i$ == `cause'}
            \STATE prediction $\leftarrow$ $\arg\max\left[\Delta_{\mathcal{M}}(c_1^i, p^i), \Delta_{\mathcal{M}}(c_2^i, p^i)\right]$
        \ELSIF{$q^i$ == `effect'}
            \STATE prediction $\leftarrow$ $\arg\max\left[\Delta_{\mathcal{M}}(p^i, c_1^i), \Delta_{\mathcal{M}}(p^i, c_2^i)\right]$
        \ENDIF
        \STATE $\eta$ $\leftarrow$ $\eta + \mathbbm{1}(\textit{prediction} == \textit{l}_i)$
    \ENDFOR
    \RETURN $\eta / |\mathcal{D}|$
\end{algorithmic}
\end{algorithm}

\begin{algorithm}[t]
\small
\caption{Computing Causal Estimand}\label{alg:causal_estimand_e1_e2}
\begin{algorithmic}
    \STATE $E_1,E_2$ : Events in an given activity
    \STATE $\mathcal{T}_n$ : Set of $n$ trajectories (temporally ordered sequence of events)\\
    \STATE $s_{\mathcal{M}}(E_1,E_2,\phi)$ $\leftarrow$  score of token \textbf{A} (associated with option `$increase$') using prompt $\phi$ of model $\mathcal{M}$.\\
    
     \STATE $s_{\mathcal{M}}(E_1,E_2,\phi_f)$ $\leftarrow$  score of token $\textbf{B}$ (associated with option `$increase$') in prompt $\phi_f$ (flipped options in prompt $\phi$) of model $\mathcal{M}$.\\
        
      \STATE $\tilde{s}_{\mathcal{M}}(E_1,E_2,\phi)$ $\leftarrow$  score of token \textbf{B} (associated with option `$decrease$') using prompt $\phi$ of model $\mathcal{M}$.\\
     \STATE $\tilde{s}_{\mathcal{M}}(E_1,E_2,\phi_f)$ $\leftarrow$  score of token \textbf{A} (associated with option `$decrease$') in prompt $\phi_f$ (flipped options in prompt $\phi$)of model $\mathcal{M}$\\
     
     \STATE Norm. Score $\displaystyle f_{\mathcal{M}}(E_1,E_2,\phi)\leftarrow\frac{s_{\mathcal{M}}(E_1,E_2,\phi)+s_{\mathcal{M}}(E_1,E_2,\phi_f)}{s_{\mathcal{M}}(E_1,E_2,\phi)+s_{\mathcal{M}}(E_1,E_2,\phi_f)+\tilde{s}_{\mathcal{M}}(E_1,E_2,\phi)+\tilde{s}_{\mathcal{M}}(E_1,E_2,\phi_f)}$\\ 

    \STATE $\hat{p}(E_2 |E_1 , z = t) \leftarrow f_{\mathcal{M}}((t,E_1),E_2,\phi)$ \\
    \STATE $\hat{p}(E_2 |\neg E_1 , z = t) \leftarrow f_{\mathcal{M}}((t,\neg E_1),E_2,\phi)$\\


            \STATE $\displaystyle p_{\mathcal{M}} (E_2 | do (E_1)) $$\leftarrow$ $\frac{1}{|\mathcal{T}_n|}\sum_{t \in \mathcal{T}_n} \hat{p} (E_2 | E_1 , z = t)$ \\
              \STATE $ \displaystyle p_{\mathcal{M}} (E_2 | do (\neg E_1)) $$\leftarrow$ $\frac{1}{|\mathcal{T}_n|}\sum_{t \in \mathcal{T}_n} \hat{p} (E_2 | \neg E_1 , z = t)$ \\
         \STATE $\displaystyle \Delta_{\mathcal{M}} \leftarrow p_{\mathcal{M}} (E_2 | do (E_1)) - p_{\mathcal{M}} (E_2 | do (\neg E_1))$ \\
    \RETURN $\Delta_{\mathcal{M}}$
\end{algorithmic}
\end{algorithm}

\section{Backdoor Adjustments}\label{app-sec:backdoor}
A set of variables $W$ satisfies the backdoor criterion relative to $T$ and $Y$ if the following are true 
\begin{itemize}
    \item[(A)] $W$ blocks all backdoor paths from $T$ to $Y$ i.e. blocking confounding or non-causation association paths
    \item[(B)] $W$ doesn’t contain any descendants of $T$
\end{itemize}

Then, $W$ satisfies the backdoor criterion \citep{pearl2016causal,neal2020introduction}.

Adhering to the above conditions of the backdoor criterion, it is reasonable to assume that the trajectory $t$ (temporally ordered sequence of events) till $E_1$ will contain the events that confound the event $E_1$ and event $E_2$ (condition A). Every event trajectory till $E_1$ will temporally precede $E_1$(condition B).
Hence, the trajectory variable will satisfy backdoor criteria in the proposed closed system. The domain of the trajectory variable is a set of all trajectories till $E_1$. 
 Therefore, conditioning on $t$ closes all paths that induce non-causal associations.
The generic representation of an approximate causal graphical model involving $E_1, E_2$, and $t$ is shown in Figure \ref{fig:causal_model_backdoor}, and can be formulated as:

\begin{equation*}
    \displaystyle p_{\mathcal{M}} (E_2 | do (E_1)) = \frac{1}{|\mathcal{T}|}\sum_{t \in \mathcal{T}} \hat{p} (E_2 | E_1 , z = t)
\end{equation*}

Where $\mathcal{T}$ is a set of all trajectories from the start of the activity till the event $E_1$(excluding $E_1$).

\section{Experiments and Results} \label{app-sec:results}
\subsection{Compute Resources} \label{app-sec:compute}
We perform all the experiments using a machine with 5 NVIDIA A100 GPUs. We use only the open-weight models with frozen parameters to present the results for better reproducibility in the future.
\subsection{Evaluation using Average Treatment Effect (ATE)} \label{app-sec:ate-results}
\noindent\textbf{Establishing Causal Relationships:} 
To validate the causal reasoning ability, the MCQA-based approach can be further extended to estimate the causal estimation and denote the causal strengths between the events. Establishing cause-and-effect relationships can be achieved through various statistical analyses. The strength of cause-and-effect relationships is approximated by statistically analyzing events' behavior using observational data (PC-Algorithm, \citet{spirtes2000causation}). Moreover, some of the recent works \citep{wang-etal-2023-cola} highlight the role of context in determining the causal relationships between the events. To extend our analysis of causal reasoning abilities in the proposed framework, we use the backdoor adjustments in LLMs as explained in the main paper. Moreover, we also perform an interesting analysis of the observational graphs for estimating $\Delta$ statistically.

\noindent\textbf{1) Through Original Trajectories:}
DeScript \cite{wanzare-etal-2016-crowdsourced} collects data by considering $\sim100$ ESDs written by different crowd-sourced workers. 
We use the original Trajectories (ESDs) $\mathcal{T}_o$ written by humans present in the DeScript dataset.
These ESDs provide the original flow in the graph directly coming from crowdsourced workers. We consider these as the original trajectories $\mathcal{T}_o$. 
Applying the backdoor criterion (Eq. \ref{eq:backdoor}) over these trajectories $\mathcal{T}_o$. An interventional distribution similar to the previous section is computed considering the likelihood of occurrence of $E_2$ under each treatment ($E_1$ and $\neg E_1$) for only these trajectories $\mathcal{T}_o$. These estimations are further used to compute the treatment effect using the Eq.  \ref{eq:average_treatment_effect}. We denote the causal risk difference ($\Delta$) computed with $\mathcal{T}_o$ as $\Delta_o$.

\noindent\textbf{2) Through Observational Graphs:}
The observational graphs provide a proxy for the underlying knowledge about the activity, covering all possible sets of events, i.e., starting from the start node, one can trace multiple trajectories that will essentially define the way of performing the activity. For every pair of connected events $(e_i, e_j)$, the edge between them represents the probable transition from $e_i$ to $e_j$ with some non-zero probability. 
However, a noteworthy point is that the transition probability between two connected events $(e_i, e_j)$ can vary depending on the design choices/transition function $T(e_i, e_j) \rightarrow (0, 1]$. 
We define this transition function in two ways:
1) \textbf{Uniform Node Transition} ($T_n$): The transition probability from current node $e_i$ to next probable events $e_j \in E_{ij}$ would be uniform that is $T (e_i,e_j) = \frac{1}{|o_i|}$, where $|o_i|$ represents number of outgoing edges from event $e_i$. (i.e., assuming after an event, the choice of the next event is uniform from the possible events).
2) \textbf{Uniform Trajectory Transition ($T_t$):} Another way to take the set of events in an activity (trajectory) is by considering all the possible paths being equally probable, i.e., across the entire population the same activity will be represented with one of the possible trajectories. Hence we can define the transition function with each trajectory $t_i= (e_{start},e_2,...,e_{end})$ (sequence of events from starting to ending) having the same probability, i.e.:
    \begin{align*}
        p (t_i) &= p (t_j) \forall t_i , t_j \in \mathcal{T} \\
    p (t_i) &= \prod_{l,m \in t_i} T_t (e_l \rightarrow e_m)
    \end{align*}

\noindent Further, given a transition function $T$, computation of $\hat{p}(E_2|E_1, z=t_i)$ becomes straightforward as $p(E_2|E_1)$, since the next course of trajectories after $E_1$ will be decided given $E_1$ has occurred. Analytically, it can be computed by counting every trajectory (i.e., $\mathcal{T}_{ij}$) from $E_1$ that leads to $E_2$ 
\begin{align}
    p (E_2|E_1) &= \sum_{t \in \mathcal{T}_{ij}} \prod_{l,m \in t} T (e_l \rightarrow e_m) \\
    p (E_2|E_1) &= \left[\sum_{k=1}^{k=M} T^k \right] (E_1 \rightarrow E_2)
    \label{eq:emperical_p_e2_given_e1}
\end{align}

\noindent For estimating the probability $\hat{p}(E_2|\neg E_1, z=t_i)$, we make use of the observational graph by considering all the parent nodes of $E_1$ and compute the probability of reaching $E_2$ from the parent (i.e., last event of trajectory $t_i$) avoiding the occurrence of $E_1$.
\setlength{\abovedisplayskip}{2pt}
\setlength{\belowdisplayskip}{2pt}
\begin{align}
\hat{p} (E_2| \neg E_1 , z=t_i) = \sum_{\substack{t \in \mathcal{T}_{ij} \\ E_1 \notin t}} \prod_{l,m \in t} T (e_l \rightarrow e_m)    
\label{eq:emperical_p_e2_given_not_e1}
\end{align}

\noindent also, $p (z=t_i)$ can be computed as product of each transition in $t_i$, i.e 
\begin{align}
    p (z=t_i) = \prod_{ij} T (e_i\rightarrow e_j)
    \label{eq:emperical_p_z_trajectory}
\end{align}

\noindent Computations from Equations \ref{eq:emperical_p_e2_given_e1}, \ref{eq:emperical_p_e2_given_not_e1}, and \ref{eq:emperical_p_z_trajectory} are used in the backdoor adjustment defined in Equation \ref{eq:backdoor} for estimating the interventional likelihood of occurrence of $E_2$ under each treatment ($E_1$ and $\neg 
 E_1$) and causal risk difference ($\Delta$). Note depending on the choice of transition function ($T_n$ or $T_t$), we obtain two deltas $\Delta_n$ and $\Delta_t$, respectively. 


\section{Prompt Templates for Language Model based Experiments} \label{app-sec:prompts}

We present the various prompt templates used to estimate the temporal link between the events in Figure \ref{fig:prompt_template_before_after}. For BERT-based models, we use the MLM-trained models for predicting the masked token given a sentence (Previously, a similar approach was adopted by \cite{pmlr-v162-zhang22am-rock}). In contrast, for autoregressive models, we frame the prompt as a question-answer objective, taking inspiration from \citep{robinson2023leveraging}, where a multiple-choice-based question is framed to predict the answer in the form of the option IDs. The prompt is intentionally structured so that the LLM is intended to predict a single token (Such as ``A'', ``B'', etc.). \citet{robinson2023leveraging}  highlights the advantages of MCQA-based evaluation over cloze evaluation (where the LLMs are expected to generate the entire answer in a cloze test), leading to a significant boost in various tasks, including commonsense-based tasks. 

For our prompt-based evaluation experiments over the generated causal triplets, we follow the same MCQA-based strategy and frame the prompts accordingly for a fair evaluation. Figure \ref{fig:prompt_template_autoregressive} presents various prompt templates for autoregressive experiments, and Figure \ref{fig:prompt_template_autoregressive_qualitative} shows a few qualitative examples for the framed causal query templates.

\begin{figure}[ht]
\centering
    \scalebox{0.85}{
    \begin{tabular}{p{1.1\linewidth}}
      \toprule
      \texttt{Consider the activity of \textcolor{orange}{activity name}.}
      \\
      \texttt{\textcolor{blue}{[ in-context examples (if few-shot/in-context learning experiment) ]}} \\
      \texttt{Which of the following events (given as options A or B) is a plausible \textcolor{orange}{question (cause/effect)} of the event \textcolor{orange}{premise}?
      }
      \\
      \texttt{A. \textcolor{blue}{\textcolor{orange}{choice1}}} \\
      \texttt{B. \textcolor{blue}{\textcolor{orange}{choice2}}} \\
      \texttt{Answer:\textcolor{red}{ \underline{ A}}} \\
      \bottomrule
    \end{tabular}
    }
    \scalebox{0.85}{
    \begin{tabular}{p{1.1\linewidth}}
      \texttt{The following are multiple choice questions about \textcolor{orange}{activity name}. You should directly answer the question by choosing the correct option.}
      \\
      \texttt{\textcolor{blue}{[ in-context examples (if few-shot/in-context learning experiment) ]}} \\
      \texttt{Which of the following events (given as options A or B) is a plausible \textcolor{orange}{question (cause/effect)} of the event \textcolor{orange}{premise}?
      }
      \\
      \texttt{A. \textcolor{blue}{\textcolor{orange}{choice1}}} \\
      \texttt{B. \textcolor{blue}{\textcolor{orange}{choice2}}} \\
      \texttt{Answer:\textcolor{red}{ \underline{ A}}} \\
      \bottomrule
    \end{tabular}
    }
\caption[Input prompt formats for the MCQA-based evaluation of autoregressive open-weight models]{
Input prompt formats for the MCQA-based evaluation of autoregressive open-weight models (e.g., \texttt{llama(-2)}, \texttt{GPT-J}, etc.).
The \texttt{black text} is the templated input. The \texttt{\textcolor{orange}{orange text}} is the input from the created causal query triplets, where the \texttt{\textcolor{orange}{activity name}} denotes the description of the activity like \texttt{baking a cake}.
The next-token prediction probabilities of the option IDs at the \textcolor{red}{\underline{\texttt{red text}}} is used as the observed prediction distribution. 
}
\label{fig:prompt_template_autoregressive}
\end{figure}

\begin{figure}[ht]
\centering
    \scalebox{0.85}{
    \begin{tabular}{p{1.1\linewidth}}
      \toprule
      \texttt{Consider the activity of \textcolor{orange}{baking a cake}.}
      \\
      \texttt{Which of the following events (given as options A or B) is a plausible \textcolor{orange}{effect} of the event \textcolor{orange}{preheat oven to 350 degrees.}?
      }
      \\
      \texttt{A. \textcolor{blue}{\textcolor{orange}{turn off oven.}}} \\
      \texttt{B. \textcolor{blue}{\textcolor{orange}{prepare the microwave oven and required utensils}}} \\
      \texttt{Answer:\textcolor{red}{ \underline{ A}}} \\
      \bottomrule
    \end{tabular}
    }
    \scalebox{0.85}{
    \begin{tabular}{p{1.1\linewidth}}
      \texttt{The following are multiple choice questions about \textcolor{orange}{going on a train}. You should directly answer the question by choosing the correct option.}
      \\
      \texttt{Which of the following events (given as options A or B) is a plausible \textcolor{orange}{cause} of the event \textcolor{orange}{get the bill for groceries.}?
      }
      \\
      \texttt{A. \textcolor{blue}{\textcolor{orange}{pay the cashier for your items.}}} \\
      \texttt{B. \textcolor{blue}{\textcolor{orange}{place cart into cart corral.}}} \\
      \texttt{Answer:\textcolor{red}{ \underline{ A}}} \\
      \bottomrule
    \end{tabular}
    }

\caption{
Qualitative examples for the MCQA-based evaluation of autoregressive open-weight models (e.g., \texttt{llama(-2)}, \texttt{GPT-J}, etc.).
}
\label{fig:prompt_template_autoregressive_qualitative}
\end{figure}

\begin{figure}[ht]
\centering
    \scalebox{0.85}{
    \begin{tabular}{p{1.1\linewidth}}
      \toprule
      \texttt{In terms of 'before' and 'after', the event: ``\textcolor{orange}{first event text}" would have happened 
        \textcolor{red}{$<$mask\_token$>$}
         the event: ``\textcolor{orange}{second event text}"} \\
      \bottomrule
    \end{tabular}
    }
    \scalebox{0.85}{
    \begin{tabular}{p{1.1\linewidth}}
      \texttt{Consider the activity of \textcolor{orange}{activity name}.}
      \\
      \texttt{Question: Determine the temporal order.} \\ 
      \texttt{The following events took place: 1. \textcolor{orange}{first event text}, 2. \textcolor{orange}{second event text}} \\ 
      \texttt{Did the first event occur 'before' or 'after' the second event? (choose from the given options)} \\
      \texttt{A: before} \\ 
      \texttt{B: after} \\ 
      \texttt{Answer:\textcolor{red}{ \underline{ A}}} \\
      \bottomrule
    \end{tabular}
    }
    
\caption[Input prompt formats for the $\Delta$ estimation (temporal) via Language models.]{
Input prompt formats for the $\Delta$ estimation via Language models. The first row shows the prompt template used for BERT-based language models, where the mask token is predicted. The second row shows the template for autoregressive open-weight models (e.g., \texttt{llama(-2)}, \texttt{GPT-J}, etc.).
The \texttt{black text} is the templated input. The \texttt{\textcolor{orange}{orange text}} is the input from the created causal query triplets, where the \texttt{\textcolor{orange}{first event text}} and \texttt{\textcolor{orange}{second event text}} comes from the premise and available set of choices.
The mask-token prediction probabilities of `before' and `after' and 
next-token prediction probabilities of the option IDs at the \textcolor{red}{\underline{\texttt{red text}}} are used as the observed prediction distribution for BERT-based and GPT-based open-weight models.
}
\label{fig:prompt_template_before_after}
\end{figure}

\begin{figure}[ht]
\centering
    \scalebox{0.85}{
    \begin{tabular}{p{1.1\linewidth}}
      \toprule
      \texttt{CAUSAL REASONING ANALYSIS:}\\
      \texttt{Context: For the activity \textcolor{orange}{activity name}.}
      \texttt{During the activity, the following set of sequences occurred in order:}
      \\
      \texttt{\textcolor{orange}{[ ordered list of events present in a Trajectory till $E_1$]}} \hfill {\textcolor{blue}{{\#\  Trajectory}  ($z=\mathcal{T}_i$)}}\\
      \texttt{Further, the event `\textcolor{orange}{event text for $E_1$}' took place.}  \hfill {\textcolor{blue}{{\#\  {Intervention}}  ($do( E_1)$)}}\\
      
      \texttt{Question: Given the above information, will the chances of the occurrence of the event `\textcolor{orange}{event text for $E_2$}'} increase or decrease?\\
      \texttt{A. \textcolor{blue}{\textcolor{orange}{Increase}}} \\
      \texttt{B. \textcolor{blue}{\textcolor{orange}{Decrease}}} \\
      \texttt{Answer:\textcolor{red}{ \underline{ A}}}  \hfill {\textcolor{blue}{\textbf{\#\ }  ($p(E_2 \mid do( E_1), z=\mathcal{T}_i$)}}\\
      \bottomrule
      
    \end{tabular}
    }
    \scalebox{0.85}{
    \begin{tabular}{p{1.1\linewidth}}
      \toprule
      \texttt{CAUSAL REASONING ANALYSIS:}\\
      \texttt{Context: For the activity \textcolor{orange}{activity name}.}
      \texttt{During the activity, the following set of sequences occurred in order:}
      \\
      \texttt{\textcolor{orange}{[ ordered list of events present in a Trajectory till $E_1$]}} \hfill {\textcolor{blue}{{\#\  {Trajectory}}  ($\mathcal{T}_i$)}}\\
      \texttt{Further, the event `\textcolor{orange}{event text for $E_1$}' did NOT take place.}  \hfill {\textcolor{blue}{{\#\  {Intervention}}  ($do(\neg E_1)$)}}\\
      \texttt{Given the above information, will the chances of the occurrence of the event `\textcolor{orange}{event text for $E_2$}' increase or decrease?}\\
      \texttt{A. \textcolor{blue}{\textcolor{orange}{Increase}}} \\
      \texttt{B. \textcolor{blue}{\textcolor{orange}{Decrease}}} \\
      \texttt{Answer:\textcolor{red}{ \underline{ B}}} \hfill {\textcolor{blue}{\textbf{\#\ }  ($p(E_2 \mid do(\neg E_1), z=\mathcal{T}_i$)}}\\
      \bottomrule
    \end{tabular}
    }
    Flipped options variant of the above Prompt Template
    \scalebox{0.85}{
    \begin{tabular}{p{1.1\linewidth}}
      \toprule
      \texttt{CAUSAL REASONING ANALYSIS:}\\
      \texttt{Context: For the activity \textcolor{orange}{activity name}.}
      \texttt{During the activity, the following set of sequences occurred in order:}
      \\
      \texttt{\textcolor{orange}{[ ordered list of events present in a Trajectory till $E_1$]}} \hfill {\textcolor{blue}{{\#\  Trajectory}  ($z=\mathcal{T}_i$)}}\\
      \texttt{Further, the event `\textcolor{orange}{event text for $E_1$}' took place.}  \hfill {\textcolor{blue}{{\#\  {Intervention}}  ($do( E_1)$)}}\\
      
      \texttt{Given the above information, will the chances of the occurrence of the event `\textcolor{orange}{event text for $E_2$}'} increase or decrease?\\
      \texttt{A. \textcolor{blue}{\textcolor{orange}{Decrease}}} \\
      \texttt{B. \textcolor{blue}{\textcolor{orange}{Increase}}} \\
      \texttt{Answer:\textcolor{red}{ \underline{ B}}}  \hfill {\textcolor{blue}{\textbf{\#\ }  ($p(E_2 \mid do(E_1), z=\mathcal{T}_i$)}}\\
      \bottomrule
    \end{tabular}
    }
    \scalebox{0.85}{
    \begin{tabular}{p{1.1\linewidth}}
      \texttt{CAUSAL REASONING ANALYSIS:}\\
      \texttt{Context: For the activity \textcolor{orange}{activity name}.}
      \texttt{During the activity, the following set of sequences occurred in order:}
      \\
      \texttt{\textcolor{orange}{[ ordered list of events present in a Trajectory till $E_1$]}} \hfill {\textcolor{blue}{{\#\  {Trajectory}}  ($\mathcal{T}_i$)}}\\
      \texttt{Further, the event `\textcolor{orange}{event text for $E_1$}' did NOT take place.}  \hfill {\textcolor{blue}{{\#\  {Intervention}}  ($do(\neg E_1)$)}}\\
      \texttt{Given the above information, will the chances of the occurrence of the event `\textcolor{orange}{event text for $E_2$}' increase or decrease?}\\
      \texttt{A. \textcolor{blue}{\textcolor{orange}{Decrease}}} \\
      \texttt{B. \textcolor{blue}{\textcolor{orange}{Increase}}} \\
      \texttt{Answer:\textcolor{red}{ \underline{ A}}} \hfill {\textcolor{blue}{\textbf{\#\ }  ($p(E_2 \mid do(\neg E_1), z=\mathcal{T}_i$)}}\\
      \bottomrule
    \end{tabular}
    }
\caption[Input prompt formats for computing the causal estimand using the autoregressive open-weight models (using backdoor criterion)]{
Input prompt formats for computing the causal estimand using the autoregressive open-weight models (using backdoor criterion) (e.g., \texttt{llama(-2)}, \texttt{GPT-J}, etc.).
The \texttt{black text} is the templated input. The \texttt{\textcolor{orange}{orange text}} is the input from the created causal query triplets, where the \texttt{\textcolor{orange}{activity name}} denotes the description of the activity like \texttt{baking a cake}. The trajectory $\mathcal{T}_i$ is obtained using the observational graph $\mathcal{G}_o$ and contains the sequence of events before the event $E_1$. 
The next-token prediction probabilities of the option IDs at the \textcolor{red}{\underline{\texttt{red text}}} is used as the observed prediction distribution. 
The flipped options variants of the prompts contain the same query with flipped options (i.e., the option `\texttt{Increase}' becomes `\texttt{Decrease}' and vice versa). This is done to make the causal estimand unbiased towards the predicted option token as highlighted in Algorithm \ref{alg:causal_estimand_e1_e2}.
}
\label{fig:prompt_template_intervention}
\end{figure}


\section{Additional Results}

\subsection{Temporally Plausible Choices in Causal Triplets} \label{app-sec:temporal-results}

Some of the initial studies \citep{do-etal-2011-minimally} highlight the difficulty in choosing between the causal-effect events and temporal events (that occur in close proximity to the premise event), i.e., temporal relationships are sometimes considered as a causal relationship by human annotators. We also create another version of created causal triplets where the wrong choices are replaced by temporally near nodes (nodes that are at a one-hop distance from the premise node). We call these `\textit{causally hard triplets}.' Note the temporal nodes are obtained from the observational graphs $\mathcal{G}_o$. Table \ref{app:tab:LLM_evaluation_results} shows the performance comparison with causal triplets and causal-temporal triplets versions of the same queries. We observe a significant performance drop on the causal-temporal triplets version for most models, highlighting the increased confusion.


\begin{table*}[t]
\caption[Results for Causal and Causal Temporal Triplets]{The table provides evaluation results of Language models over the causal and causal temporal triplets.}
\label{app:tab:LLM_evaluation_results}
\centering
\small
\renewcommand{\arraystretch}{1}
\setlength\tabcolsep{5pt}
{%
\begin{tabular}{clccccc}
\toprule
\textbf{Triplets}                                  & \textbf{Model Name} & \texttt{\textbf{cake}} & \texttt{\textbf{shopping}} & \texttt{\textbf{train}} & \texttt{\textbf{tree}} & \texttt{\textbf{bus}} \\ 
\midrule
\multirow{10}{*}{\rotatebox[origin=c]{90}{causal triplets}}  & gpt-neo-125M            & \textcolor{blue}{50.71} & \textcolor{blue}{50.01} & 49.99 & \textcolor{blue}{50.13} & 50.15         \\
                                                       & gpt-neo-1.3B     &    44.77 & 45.69 & 42.52 & 45.67 & 42.89        \\
                                                       & gemma-2b        & \textcolor{blue}{53.76} & \textcolor{blue}{52.19} & \textcolor{blue}{60.57} & \textcolor{blue}{60.71 } & \textcolor{blue}{53.64}        \\
                                                       & gpt-neo-2.7B  & 50.00 & 50.01 & 50.00 & 50.01 & 50.00        \\
                                                       & phi-2     & \textcolor{blue}{85.14} & \textcolor{blue}{83.65} & \textcolor{blue}{77.29} & 82.24 & \textcolor{blue}{71.74}        \\
                                                       & gpt-j-6B     & 49.59 & 50.02 & 50.29 & 49.92 & 49.93         \\
                                                       & Llama-2-7b-chat-hf               & \textcolor{blue}{77.92} & \textcolor{blue}{72.41} & \textcolor{blue}{73.48} & 72.40 & \textcolor{blue}{68.21}        \\
                                                       & Mistral-7B-v0.1 & \textcolor{blue}{77.64} & \textcolor{blue}{69.38} & \textcolor{blue}{68.46} & \textcolor{blue}{72.43} & \textcolor{blue}{69.37}        \\ 
                                                       & gemma-7b
                                                       & \textcolor{blue}{81.47} & \textcolor{blue}{82.26} & \textcolor{blue}{77.24} & 80.78 & \textcolor{blue}{70.29}       \\ 
                                                       & Meta-Llama-3-8B
                                                       & \textcolor{blue}{80.79} & \textcolor{blue}{76.46} & \textcolor{blue}{76.08} & 78.21 & \textcolor{blue}{67.39}        \\
\midrule
\multirow{10}{*}{\rotatebox[origin=c]{90}{causally hard triplets}}  & gpt-neo-125M            & 50.60 & 49.80 & 49.90 & 50.00 & 50.20         \\
                                                       & gpt-neo-1.3B     &    49.50 & 51.20 & 48.80 & 47.50 & 48.00     \\
                                                       & gemma-2b        &  52.30 & 51.00 & 56.10 & 52.20 & 50.00        \\
                                                       & gpt-neo-2.7B  & 50.00 & 50.00 & 50.00 & 50.00 & 50.00     \\
                                                       & phi-2     & 80.00 & 74.70 & 67.90 & 87.50 & 66.50        \\
                                                       & gpt-j-6B     & 50.20 & 50.00 & 50.30 & 50.00 & 49.80       \\
                                                       & Llama-2-7b-chat-hf               & 71.60 & 66.60 & 68.30 & 77.00 & 65.40     \\
                                                       & Mistral-7B-v0.1 &  69.20 & 63.10 & 64.20 & 67.90 & 62.00      \\ 
                                                       & gemma-7b
                                                       &  76.30 & 76.40 & 69.70 & 89.70 & 63.70       \\ 
                                                       & Meta-Llama-3-8B
                                                       & 77.30 & 72.20 & 69.30 & 83.20 & 64.30      \\
                                                       \bottomrule
\end{tabular}}
\end{table*}

